%% file: main.tex
\documentclass{article}

    \PassOptionsToPackage{numbers, compress}{natbib}
    \usepackage[final]{neurips_2020}

\usepackage[utf8]{inputenc} % allow utf-8 input
\usepackage[T1]{fontenc}    % use 8-bit T1 fonts
\usepackage[colorlinks,linktoc=all]{hyperref}       % hyperlinks
\usepackage{url}            % simple URL typesetting
\usepackage{booktabs}       % professional-quality tables
\usepackage{amsfonts}       % blackboard math symbols
\usepackage{nicefrac}       % compact symbols for 1/2, etc.
\usepackage{microtype}      % microtypography

\usepackage{graphicx}
\usepackage{booktabs} % for professional tables
\usepackage[usenames,dvipsnames]{xcolor}
\usepackage[algo2e,ruled,linesnumbered,lined]{algorithm2e}
\usepackage{amsmath}
\usepackage{amsfonts}

\usepackage{makecell}
\usepackage{soul}
\usepackage{wrapfig}
\usepackage{caption}
\usepackage{lipsum}
\usepackage{subcaption}
\usepackage[colorinlistoftodos,disable]{todonotes}
\usepackage{comment}
\usepackage{bm}

\hypersetup{citecolor=BurntOrange}
\hypersetup{linkcolor=MidnightBlue}
\hypersetup{urlcolor=MidnightBlue}
\usepackage[capitalise,nameinlink]{cleveref}
\crefname{algorithm}{Alg.}{Algs.}
\Crefname{algorithm}{Algorithm}{Algorithms}
\crefname{section}{Sec.}{Secs.}
\Crefname{section}{Section}{Sections}
\crefname{table}{Tab.}{Tabs.}
\Crefname{table}{Table}{Tables}
\crefname{appendix}{App.}{Apps.}
\Crefname{appendix}{Appendix}{Appendixes}

\newcommand{\std}[1]{ \color{darkgray}\tiny{$\pm$#1} }

\renewcommand{\paragraph}[1]{{\bf #1}\;\;\;}

\newcommand{\newsetting}{Online faSt Adaptation and Knowledge Accumulation} 
\newcommand{\newacron}{OSAKA}

\newcommand{\firstFeature}{update modulation}
\newcommand{\FirstFeature}{Update modulation}
\newcommand{\firstFeatureAcron}{UM} 

\newcommand{\secondFeature}{prolonged adaptation phase} 
\newcommand{\SecondFeature}{Prolonged adaptation phase} 
\newcommand{\secondFeatureAcron}{PAP}

\newcommand\blfootnote[1]{%
  \begingroup
  \renewcommand\thefootnote{}\footnote{#1}%
  \addtocounter{footnote}{-1}%
  \endgroup
}

\title{Online Fast Adaptation and Knowledge Accumulation (OSAKA): \\ a New Approach to Continual Learning}
\author{%
    Massimo Caccia$^{123}$ \quad Pau Rodr\'iguez$^2$ \quad Oleksiy Ostapenko$^{13}$ \quad Fabrice Normandin$^{13}$ \\
    \bf Min Lin$^{13}$  \quad Lucas Caccia$^{145}$  \quad Issam Laradji$^2$  \quad Irina Rish$^{137}$ \\
    \bf Alexandre Lacoste$^2$  \quad David Vazquez$^2$  \quad Laurent Charlin$^{167}$ \\
    \\
    $^1$Mila - Quebec AI Institute, $^2$ElementAI, $^3$Université de Montréal, $^4$Facebook AI Research\\
    $^5$McGill University, $^6$HEC Montréal,  $^7$Canada CIFAR AI Chair \\
}

\begin{document}

\maketitle

%------------------------------------------- ABSTRACT ------------------------------------------------------%
%-----------------------------------------------------------------------------------------------------------%
\begin{abstract}

Continual learning agents experience a stream of (related) tasks. The main challenge is that the agent must not forget previous tasks and also adapt to novel tasks in the stream. We are interested in the intersection of two recent continual-learning scenarios. In \emph{meta-continual learning}, the model is pre-trained using meta-learning to minimize catastrophic forgetting of previous tasks. In \emph{continual-meta learning}, the aim is to train agents for \emph{faster remembering} of previous tasks through adaptation. In their original formulations, both methods have limitations. We stand on their shoulders to propose a more general scenario, OSAKA, where an agent must quickly solve new (out-of-distribution) tasks, while also requiring fast remembering. We show that current continual learning, meta-learning, meta-continual learning, and continual-meta learning techniques fail in this new scenario.
We propose \emph{Continual-MAML}, an online extension of the popular MAML algorithm as a strong baseline for this scenario. We show in an empirical study that \emph{Continual-MAML} is better suited to the new scenario than the aforementioned methodologies including standard continual learning and meta-learning approaches.
 \blfootnote{corresponding author: massimo.p.caccia@gmail.com}

\end{abstract}

%------------------------------------------- INTRODUCTION -------------------------------------------------%
%----------------------------------------------------------------------------------------------------------%
\section{Introduction}
\label{sec:intro}

A common assumption in supervised machine learning is that the data is independently and identically distributed (i.i.d.). This assumption is violated in many practical applications handling non-stationary data distributions, including robotics, autonomous driving, conversational agents, and other real-time applications. Over the last few years, several methodologies study learning from non-i.i.d. data. We focus on \emph{continual learning} (CL), where the goal is to learn incrementally from a non-stationary data sequence involving different datasets or \emph{tasks}, while not forgetting previously acquired knowledge, a problem known as \emph{catastrophic forgetting}~\citep{mccloskey1989catastrophic}. 

We draw inspiration from autonomous systems deployed in environments that might differ from the ones they were (pre-)trained on. For instance, a robot pre-trained in a factory and deployed in homes where it will need to adapt to new domains and even solve new tasks. Or a virtual assistant can be pre-trained on historical data and then adapt to its user's needs and preferences once deployed. Further motivating applications exist in time-series forecasting including market prediction, game playing, autonomous customer service, recommendation systems, and autonomous driving. These systems must adapt online to maximize their cumulative rewards~\citep{kaelbling1991foundations, kaelbling1993learning}. As a step in that direction, we propose a task-incremental scenario (OSAKA) where previous tasks reoccur and new tasks appear. We measure the cumulative accuracy of models instead of the (more common) final accuracy to evaluate how quickly models and algorithms adapt to new tasks and remember previous ones.

\paragraph{Background} \emph{Task-incremental classification} is a common supervised CL scenario where classification datasets are presented to an online learner sequentially, one task at a time. For each task $T_t$ at iteration $t$, the data are sampled i.i.d.\ from their corresponding distribution $P_t({\bf x},{\bf y})$. In the task-incremental scenario models are evaluated by their average final performance across all tasks---after being trained on all tasks sequentially. Several families of recent CL approaches use this setting, including regularization methods~\citep{kirkpatrick2017overcoming}, data replay~\citep{soltoggio2015short}, and dynamic architectures~\citep{rusu2016progressive}  (see \citet{lange2019continual} and \citet{Parisi18review} for comprehensive overviews).

More recent approaches propose relaxing some constraints associated with task-incremental CL by combining CL and meta-learning. \emph{Continual-meta learning} (CML) focuses on \emph{fast remembering} or how quickly the model recovers its original performance on past tasks~\citep{He2019TaskAC}. \emph{Meta-continual learning} (MCL) uses meta-learning to learn not to forget~\citep{Javed2019Meta}. In this paper, we further extend the task-incremental setting and show empirical benefits compared to CML and MCL (see \Cref{sec:experiments}).

\paragraph{\newacron{}} We propose a more flexible and general scenario inspired by a pre-trained agent that must keep on learning new tasks after deployment. In this scenario, we are interested in the cumulative performance of the agent throughout its lifetime~\cite{kaelbling1991foundations, kaelbling1993learning}.  (Standard CL reports the final performance of the agent on all tasks at the end of its \emph{life}.) To succeed in this scenario, agents need the ability to learn new tasks as well as quickly remember old ones.

We name our CL setting  \textit{\newsetting} (\newacron). The main characteristics of \newacron{} are that at deployment or CL time: 1) task shifts are sampled stochastically, 2) the task boundaries are unknown ({\em task-agnostic} setting), 3) the target distribution is context-dependent, 4) multiple levels of non-stationarity are used, and 5) tasks can be revisited. Furthermore, our evaluation of CL performance is different from the one commonly used in CL. We report the cumulative or online average performance instead of the final performance on all seen tasks. 

Existing CL methods are not well-suited to \newacron. Methods such as EWC~\citep{kirkpatrick2017overcoming}, progressive networks~\citep{rusu2016progressive} or MCL~\citep{Javed2019Meta} require task boundaries. In contrast, task-agnostic methods (e.g. \citep{Aljundi17, zeno2018task, He2019TaskAC}) optimize for the final performance of the model and so resort to mechanisms that attempt to eliminate catastrophic forgetting. The extra computations resulting from the mechanisms hinder online performance and unnecessarily increase the computational footprint of the algorithms.

To address the challenges of \newacron{}, we propose {\em Continual-MAML}, a baseline inspired by the meta-learning approach of MAML~\citep{finn2017model}. Continual-MAML is pre-trained via meta-learning. When deployed, Continual-MAML adapts the learned parameter initialization to solve new tasks. When a change in the distribution is detected, new knowledge is added into the learned initialization. As a result, Continual-MAML is more efficient and robust to distribution changes since it does not require computationally expensive optimizers like BGD~\citep{zeno2018task} or replay methods used in prior work~\citep{Chaudhry18,shin2017continual}.

Using our \newacron~scenario, we compare the performance of Continual-MAML to recent and popular approaches from continual learning, meta-learning, and continual-meta learning. Across several datasets, we observe that Continual-MAML is better suited to \newacron{} than  prior methods from the aforementioned fields and thus provides an initial strong baseline. 

To summarize, our contributions include: (1) \newacron{}, a new CL setting which is more flexible and general than previous ones. Related, we also propose a unifying scenario for discussing meta- and continual learning scenarios (\Cref{tbl:uni}); (2) the Continual-MAML algorithm, a new baseline that addresses the challenges of the \newacron{} setting; (3) extensive empirical evaluation of our proposed method%and the current literature in this new scenario
; and (4) a codebase for researchers to test their methods in the \newacron{} scenario.\footnote{\url{https://github.com/ElementAI/osaka}}

%---------------------------------------------- UNIFY ------------------------------------------------------%
%-----------------------------------------------------------------------------------------------------------%

\section{A unifying framework}
\label{sec:setup}

\newcommand{\cl}{\operatorname{CL}}
\newcommand{\ctxt}{C}
\newcommand{\loss}{\mathcal{L}}
\newcommand{\world}{\mathcal{W}}
\newcommand{\algo}{\mathcal{A}}
\newcommand{\highlight}[1]{\textcolor{blue}{#1}}

\begin{table*}[t]
    \begin{scriptsize}
    % \begin{tiny}
    %\begin{sc}
    \begin{center}
    \resizebox{0.9\linewidth}{!}{%
    \centerline{
    \setlength{\tabcolsep}{6pt}
    \renewcommand{\arraystretch}{1.7}
    \begin{tabular}{ccccc}
    \toprule
        % \hline
        & {\scriptsize Data Distribution} & {\scriptsize Model for Fast Weights} & {\scriptsize Slow Weights Updates} & {\scriptsize Evaluation}  \\
        % & {\scriptsize Data Distribution} & {\scriptsize Model for Fast weights} & {\scriptsize Model for Slow weights} & {\scriptsize Evaluation}  \\
        \midrule
        { \scriptsize Supervised Learning} & $ S, Q \sim C $ & $ f_\theta = \algo(S) $ & --- & $ \loss(f_\theta, Q) $ \\
        \midrule 
        \makecell{\scriptsize Meta-learning} & \makecell{ $\{C_i\}_{i=1}^M \sim \world^M$ \\ $S_i, Q_i \sim C_i $ }& $ f_{\theta_i} = \algo_\phi(S_i)$ & $ \makecell{\nabla_\phi \loss( f_{\theta_i}, Q_i) \\ {\scriptstyle \forall i<N}} $ & $ \sum_{i=N}^M \loss(\algo_\phi(S_i), Q_i) $\\
        %\midrule
        \makecell{\scriptsize  Continual Learning} & $S_{1:T}, Q_{1:T} \sim C_{1:T}$ & $ f_\theta = \cl(S_{1:T}) $ & --- & $\sum_t \loss ( f_\theta, Q_t) $ \\
        \midrule
        \makecell{\scriptsize  Meta-Continual Learning} &  \makecell{ $ \{C_{i,1:T}\}_{i=1}^M \sim \world^M$ \\ $ S_{i, 1:T}, Q_{i, 1:T} \sim C_{i,1:T} $} & $f_{\theta_i} = \cl_\phi(S_{i,1:T}) $ &  \makecell{ $\nabla_\phi \sum_t \loss( f_{\theta_i} , Q_{i,t} )$ \\
         ${\scriptstyle \forall i<N}$ } & $\sum_{i=N}^M\sum_t \loss (\algo_\phi(S_{i,1:T}), Q_{i,t}) $ \\
        %\midrule 
        \makecell{\scriptsize Continual-meta learning}  & $ S_{1:T}, Q_{1:T} \sim C_{1:T}$ & $f_{\theta_t} = \algo_\phi(\highlight{S_{t-1}})$& $\nabla_\phi \loss( f_{\theta_t}, S_t)$ & $ \sum_t \loss( \algo_\phi(S_t), Q_t )$\\ %$ \sum_t \loss( \algo_\phi(S_t), Q_t )$ \\
        \midrule
        \makecell{\scriptsize \newacron{}}  & $ Q_{1:T} \sim C_{1:T}$ & $f_{\theta_t} = \algo_\phi(\highlight{Q_{t-1}})$& $\nabla_\phi \loss( f_{\theta_t}, Q_t)$ &$ \sum_t \loss( f_{\theta_t}, Q_t )$ \\  
        \bottomrule
    \end{tabular}
    %}
    }}
    \end{center}
    % \end{tiny}
    \end{scriptsize}
    \caption{\textbf{A unifying framework} for different machine learning settings. Data sampling, fast weights computation and slow weights updates as well as evaluation protocol are presented with meta-learning terminology, i.e., the support set $S$ and query set $Q$. For readability, we omit OSAKA pre-training.}
    \label{tbl:uni}
\end{table*}

We introduce the concepts and accompanying notation that we will use to describe OSAKA in \Cref{sec:osaka}. These concepts provide a unifying framework---highlighted in \Cref{tbl:uni}---for expressing several important paradigms such as continual learning, meta-learning, and variants. A motivation for this framework (and so this section in our paper) is to clarify some confusion that arose from the recent interrelation of meta-learning and continual learning. Our main contribution, \newacron{}, can be understood even if the reader chooses to skip this section.

We begin by assuming a hidden context variable $C$ that determines the data distribution, e.g., the user's mood in a recommender system or an opponent's strategy in game playing. In some fields, contexts are referred to as \emph{tasks}. In the rest of the paper, we will use both terms interchangeably. We use $\world$ to denote a finite set of all possible contexts. Given $C$, data can be sampled i.i.d.\ from $p(X|C)$. Different learning paradigms can be described by specializing the distribution $P(C)$. For example, in the classical setting data are sampled i.i.d.\ from $p(X|C)P(C)$ where $C$ could represent the set of classes to be discriminated.

We use terminology from meta-learning and introduce a \textit{support set} $S$ and a \textit{query set} $Q$ to denote the meta-training and meta-test sets~\citep{vinyals2016matching}, respectively. These sets are usually composed of $n$ i.i.d.\ samples $X_i=(\bm{x}_i,\bm{y}_i)$, generated conditionally from the context $C_i$. In some paradigms, including supervised learning, the target distribution is fixed, i.e. $p(\bm{y}|\bm{x})=p(\bm{y}|\bm{x},C)$. We refer to the setting where the equality does not hold as having \textit{context-dependent targets}. We define a learning algorithm $\algo$ as a functional taking $S$ as input and returning a predictor $f_\theta$, with $\theta$ parameters describing the behavior of the predictor, i.e.\ $f_\theta = \algo(S)$. We also define a loss function $\loss(f_\theta, Q)$ to evaluate the predictor $f_\theta$ on the query set $Q$.

In {\bf meta-learning}, $C$ represents a task descriptor or task label, and both meta-training and meta-testing sets are sampled i.i.d.\ from $p(X| C)$. E.g., in $N$-shot classification, the task descriptor specifies the $N$ classes which have to be discriminated. Targets are context-dependent in this learning paradigm. Here, we focus only on the meta-learning methods that rely on episodic training.

% NOTE: there was a error here so I had to reformulate
A meta-learning algorithm $\algo_\phi$ adapts its behavior by learning the parameters $\phi$. It samples $M$ i.i.d.\ pairs of $S$ and $Q$ from a distribution over contexts $\world^M$: $\{C_i\}_{i=1}^M \sim \world^M$ and $(S_i, Q_i) \sim X_i\mid C_i$. Assuming that the learning process is differentiable, the parameters $\phi$ are learned using the gradient from the query set, $\nabla_\phi \loss( \algo_\phi(S_i), Q_i)$. Concretely, $\phi$ is first learned on the sets $(S_i, Q_i)$, where $i<N<M$ and the final evaluation of the algorithm is $ \sum_{i=N}^M \loss(\algo_\phi(S_i), Q_i) $.

In task-incremental {\bf continual learning}, the data distribution is non-stationary, and various CL scenarios arise from specific assumptions about this non-stationarity. Here we assume that data non-stationarity is caused by a hidden process $\{C_t\}_{t=1}^T$, where $C_t$ is the context at time $t$. $C$ in continual learning can be the task label, e.g., in Permuted MNIST, disjoint MNIST/CIFAR10~\cite{kirkpatrick2017overcoming}. It could also be the class label in the class-incremental setting~\cite{rebuffi2017icarl}. Both frameworks have a fixed target distribution. $\{C_t\}_{t=1}^T$ is usually assumed to be an ordered list of the tasks/classes. 

Continual learning algorithms work with a sequence of support sets, $S_{1:T}$, and a sequence of query sets, $Q_{1:T}$, obtained from a sequence of contexts, $C_{1:T}$. A continual learning algorithm $\cl$ transforms $S_{1:T}$ into a predictor $f_\theta$, i.e.\ $f_\theta = \cl(S_{1:T})$. The main difference with a conventional algorithm $\algo$ is that the support set is observed sequentially and cannot be fully stored in memory. The evaluation is then performed independently on each $Q_t$ (obtained in the same context as $S_t$): $\sum_{t=1}^T \loss( f_\theta, Q_t)$. In \cref{app:unify} we explain how the recent meta-continual learning and continual-meta learning settings fit into the unifying framework.

\section{Online FaSt Adaptation and Knowledge Accumulation (OSAKA)} 
\label{sec:osaka}

We propose \newacron{}, a new continual-learning scenario that lifts some of constraints of current task-incremental approaches~\cite{kirkpatrick2017overcoming, Javed2019Meta, Aljundi2019Online}. \newacron{} is aligned with the use case of deploying a pre-trained agent in the real world, where it is crucial for the agent to adapt quickly to new situations and even to learn new concepts when needed. In particular, \newacron{} proposes a scenario for evaluating such continually-learning agents.

To materialize such an evaluation \newacron{} combines different ideas: 1) agents start in a pre-training stage before continual-learning starts; 2) it provides a mechanism for proposing both old and new tasks to agents where the task boundaries remain unobserved to them; 3) it evaluates agents using their cumulative performance (e.g. accuracy) to measure their capacity to adapt to new tasks. 
This evaluation implicitly allows agents to forget which may enable faster and more efficient adaptation. For instance, partially forgetting an infrequent task allows the agent to re-allocate modeling capacity to tasks that are encountered more frequently.

We now describe \newacron{} using the procedural view of \cref{algo:osaka}.
\newacron{} proposes a two-stage approach where an agent $\theta_0$ starts in a \emph{pre-training phase} (\cref{algo:osaka}, L4--L8) and then moves to a \textit{deployment phase} (\cref{algo:osaka}, L10--L16) also known as continual-learning time.

\vspace{0.4cm}

% ALGORITHMS
\SetKwIF{If}{ElseIf}{Else}{if}{}{else if}{else}{end if}%
\SetKwFor{While}{while}{}{end while}%

\centerline{
\scalebox{0.85}{
    \begin{minipage}{0.56\linewidth}
        \setlength{\textfloatsep}{3mm}
        \begin{algorithm2e}[H]
        \SetAlgoLined
        \DontPrintSemicolon
        \textbf{Require:}  $P(C_\text{pre})$, $P(C_\text{cl})$: distributions of contexts\\
        \textbf{Require:}  $\alpha$: non-stationarity level \\
        \textbf{Initialize:} $\theta_0$: Model \\
        \While{ pre-training }{
        Sample a context $C \sim P(C_\text{pre})$ \\
        Sample data from context $\bm{x}, \bm{y} \sim p(\bm{x}, \bm{y} |C)$ \\
        Update $\theta_0$ with $\bm{x}, \bm{y}$
        }
        \While{ continually learning }{
            Sample current context $C_t \sim P(C_\text{cl} | C_{t-1}; \alpha)$ \\
            Sample data from context $\bm{x}_t, \bm{y}_t \sim p(\bm{x},\bm{y} | C_t)$ \\
            % Predict $\hat{Y_t} = f_{\theta_t}(X_t)$ \\
            Incur loss $\mathcal{L}\big(\theta_{t-1}(\bm{\bm{x}}_t), \bm{y}_t\big)$ \\
            % Obtain targets $Y$; Update $\theta_t$ with $\bm{x}_t,Y_t$ at discretion \\
            Update $\theta_t$ with $\bm{x}_t,\bm{y}_t$ at discretion \\
            $t \gets t+1$ \\
        } 
            \caption{OSAKA}
        \label{algo:osaka}
        \end{algorithm2e}
    \end{minipage}
    \hspace{2em}
    \begin{minipage}{0.55\linewidth}
        \setlength{\textfloatsep}{3mm}
        \begin{algorithm2e}[H]
            \SetAlgoLined
            \DontPrintSemicolon
            \textbf{Require:} $\eta, \gamma, \lambda$: learning rate, hyperparameters \\
            \setcounter{AlgoLine}{14}
            \While{ continually learning }{
                $C_t \sim P(C_\text{cl} | C_{t-1})$ \\
                $\bm{x}_t, \bm{y}_t \sim P(\bm{x},\bm{y}|C_t)$ \\
                $\mathcal{L}\big(f_{\theta_{t-1}}(\bm{x}_t), \bm{y}_t\big)$ \\
                $\tilde{\theta_t} \gets \phi - \phi_\eta \nabla_\phi \mathcal{L}\big(f_\phi(\bm{x}_t), \bm{y}_t\big)$ \\
                \uIf{$\mathcal{L}\big(f_{\theta_{t-1}}(\bm{x}_t), \bm{y}_t\big) - \mathcal{L}\big(f_{\tilde{\theta_{t}}}(\bm{x}_t), \bm{y}_t\big) < \gamma$}{
                %  \# No context shift detected \\
                $\theta_t \gets \theta_{t-1} - \phi_\eta \nabla_\theta \mathcal{L}\big(f_{\theta_{t-1}}(\bm{x}_t), \bm{y}_t\big)$ \\
                }
              \uElse{
                %  \# Task boundary detected \\
                % $\eta_t \gets  \eta \lambda  \mathcal{L}\big(f_{\theta_{i-1}}(\bm{x}_i), Y_i\big)$ \\
                $\eta_t \gets  \eta  g_\lambda \big( \mathcal{L}\big(f_{\theta_{t-2}}(\bm{x}_{t-1}), \bm{y}_{t-1}\big) \big)$ \\
                $\phi \gets \phi - \eta_t \nabla_\phi \mathcal{L}\big(f_{\theta_{t-2}}(\bm{x}_{t-1}), \bm{y}_{t-1}\big)$ \\
                $\theta_t \gets \phi - \phi_\eta \nabla_\phi \mathcal{L}\big(f_\phi(\bm{x}_t), \bm{y}_t\big)$  \\
                }
                $t \gets t+1$ \\
            }
            \caption{Continual-MAML at CL time}
        \label{algo:cmaml}
        \end{algorithm2e}
    \end{minipage}
    }
}

\vspace{0.4cm}

\paragraph{Pre-training (\cref{algo:osaka} L4--L8).}
In many current settings~\cite{kirkpatrick2017overcoming, He2019TaskAC}, the agent begins learning from randomly-initialized parameters. However, in many scenarios, it is unrealistic to deploy an agent without any world knowledge~\cite{lesort2019continual,lomonaco2019fine}, in part, since real-life non-i.i.d.\ training is difficult to learn. Further, in many domains, ample pre-training data can be leveraged.  

{\bf Continual-learning time (\cref{algo:osaka} L9--L15)} After pre-training, a stream of continual learning tasks evaluate the model. Each iteration $t$ in the stream relies on a context $C_t$ which determines the current task $(\mathbf{x}_t, \mathbf{y}_t)$. The contexts follow a Markov process $\{C_t\}_{t=1}^T$ with transition probabilities $P(C_t|C_{t-1};\alpha)$ (\cref{algo:osaka}, L10). 

The context is at the heart of \newacron{} and its process controls the level of stationarity of the continual-learning stage and it enables both revisiting tasks and out-of-distribution ones as well as context-dependent targets. We discuss these features below.

{\bf Controllable non-stationarity.}
\newacron{} provides control, through a hyperparameter, over the level of non-stationarity of the Markov chain. A stream is $\alpha$-locally-stationary when $P(C_t\!=\!c | C_{t-1}\!=\!c)\!=\!\alpha$. Namely, the data distribution is stationary within a local-time window, i.e., over a certain amount of timesteps. Control over $\alpha$ enables exploring environments with different levels of non-stationarity to test algorithmic robustness.

Similar to the few-shot learning literature \cite{vinyals2016matching,ravi2016optimization,oreshkin2018tadam,rodriguez2020embedding}, the transitions of the context variables in OSAKA are not structured, i.e. the context transition matrix that encodes the probability of transitioning from context $i$ to context $j$ has $\alpha$ on the diagonal and $(1-\alpha)/(|C|-1)$ everywhere else. For that reason, modelling the evolution of the context variables is not essential. Further, in OSAKA the environment provides enough feedback to the agents for re-adaptation via the targets $\bm{y}_t$ (\cref{algo:osaka}, L13). We leave the design of a continual learning experimental setup and associated modeling with a structured context variable for future work.

{\bf Task revisiting.} 
Standard CL methods incrementally learn strictly new tasks. However, many CL applications require revisiting previous tasks. Through the process $\{C_t\}_{t=1}^T$  \newacron{} proposes task revisiting, analogous to \emph{recurrent concept drift}~\cite{gama2014survey} in online learning. By revisiting previous data distributions, methods will enjoy the same form of implicit replay that agents and systems naturally benefit from in real-world scenarios. The domain of each context $C_t$ contains all tasks and so the process allows to switch back and forth from old tasks to OoD tasks. 

{\bf Out-of-distribution (OoD) tasks.} 
Current settings that permit pre-training then continually learn tasks sampled from the same data distribution~\cite{Javed2019Meta, beaulieu2020learning}. In contrast, in \newacron{} the model has to learn online tasks sampled from new distributions not encountered at pre-training (see \Cref{ssec:datasets} for details). This setting is more realistic since an agent will encounter unexpected situations in real life requiring the algorithm to update its representations. 

{\bf Context-dependent targets.} 
In standard CL, $p_t(\bm{x})$ shifts over time, but the target distribution $p(\bm{y}|\bm{x})$ is fixed. However, drift in the target distribution is common in multiple applications and is studied extensively in online learning as \textit{real concept drift} \cite{gama2014survey}. Extending \cite{He2019TaskAC}, OSAKA allows for context-dependent targets (\cref{algo:osaka}, L11) making it more flexible and more aligned with our use-cases (see \cref{sec:intro}). In \newacron{} the target distribution is  $p(\bm{y}|\bm{x},C_t)$.

% The context variable in \newacron{} is generic but it is motivated by real-world domains. For example, the context variable could be the strategy of an opponent in a game \cite{silver2016mastering,moravvcik2017deepstack,vinyals2019grandmaster,OpenAI_dota}, regimes in time-series forecasting \cite{rabiner1989tutorial,ghahramani2001introduction,caccia2017option} or the mood of a user when navigating a content platform in recommender systems \cite{hidasi2015session,song2019session}. In all these examples, like in \newacron{}, the targets change over time based on a context. Similarly, it can represent unobserved variables in partially observable Markov decision processes (POMDPs) \cite{kaelbling1998planning}, hidden contexts in hidden-mode Markov decision processes \citep{choi2000hidden}, or tasks' variations in robotics \cite{traore2019discorl,James2020RLBenchTR,Yu2019MetaWorldAB}.

The context variable in \newacron{} is generic but it is motivated by real-world domains. For example, the context variable could be the strategy of an opponent in a game \cite{silver2016mastering,moravvcik2017deepstack,vinyals2019grandmaster,OpenAI_dota}, regimes in time-series forecasting \cite{rabiner1989tutorial,ghahramani2001introduction,caccia2017option}, the mood of a user when navigating a content platform in recommender systems \cite{hidasi2015session,song2019session} or any unobserved variable in RL, e.g., in partially observable Markov decision processes (POMDPs) \cite{kaelbling1998planning} or in hidden-mode Markov decision processes \citep{choi2000hidden}. In all these examples, like in \newacron{}, the targets change over time based on a context.

{\bf Task agnostic.} In \newacron{}
the agent does not observe the task boundaries or context shifts, and it must infer the current task or context $C_t$. This is called task-agnostic (or task-free) CL~\cite{Aljundi2019TaskFreeCL,Aljundi2019Gradient, zeno2018task, He2019TaskAC, lesort2019continual} and is motivated by real-world scenarios where signals explicitly indicating a shift may not exist.

{\bf Online Evaluation (\cref{algo:osaka} L12).} 
Current settings reward methods that retain their performance on all previously seen tasks. This is an unrealistic constraint, particularly under limited computational resources~\cite{kaelbling1991foundations, kaelbling1993learning}. Instead of measuring the final performance of the model, \newacron{} measures the online cumulative performance which better suits non-stationary environments. Models are evaluated in an online fashion using the sum of the losses across all timesteps  $\sum_{t=1}^T{\mathcal{L}(f_{\theta_t},Q_t)}$ where $\mathcal{L}$ can be any loss (\cref{algo:osaka}, L12). This is as opposed to reporting only the final accuracy---for example  $\sum_{t=1}^T{\mathcal{L}(f_{\theta_T},Q_t)}$ \cite{kirkpatrick2017overcoming,rebuffi2017icarl,Chaudhry18,Chaudhry19,Javed2019Meta}. Similar to the final accuracy, the online cumulative accuracy measures both plasticity and stability. Specifically, plasticity is evaluated when the algorithm encounters OoD tasks requiring additional learning. Models with higher stability can recover past performance faster and thus enjoy higher online cumulative performance. The cumulative accuracy is also similar to evaluating the (undiscounted) sum of rewards in reinforcement learning or the regret~\citep{berry1985bandit} in online learning albeit without the need to compute the performance of an optimal model.

We instantiate OSAKA for image classification tasks (see \cref{sec:experiments}), similarly to the majority of CL benchmarks. Some motivations for our proposed experimental setting are drawn, however, from a reinforcement learning (RL) scenario. In fact, 
we could adapt OSAKA to RL. We could replace the image classification tasks by tasks from multi-task RL benchmarks (e.g., such as different robotic tasks \cite{Yu2019MetaWorldAB,James2020RLBenchTR}). We could also use a standard RL benchmark, e.g, from a model-based control environment \cite{todorov2012mujoco}, and create different contexts by changing some of the environment variables, e.g. the gravity. 
Once tasks or contexts are defined, we could group them in such a way that the CL-time tasks are OoD with respect to pre-training tasks. 
Increasingly more complex tasks could also be introduced at CL time to mimic a curriculum-learning scenario. 
Finally, to control for different levels of non-stationarity, we could adjust the time allocation or the number of episodes in each context/task.

%---------------------------------------------- METHODOLOGY ------------------------------------------------%
%-----------------------------------------------------------------------------------------------------------%

\section{Continual-MAML}
\label{sec:method}

\begin{figure*}[b]
    \centering
    \includegraphics[trim={2cm 1cm 1cm 1cm},clip,width=0.9\linewidth]{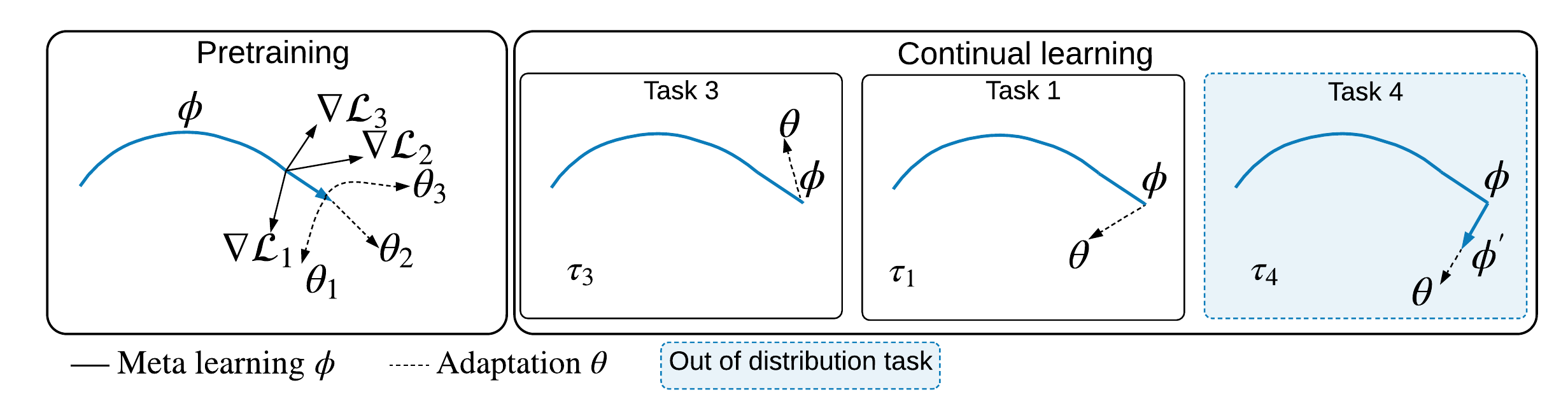}
    \label{fig:problem_formulation}
    \caption{ \textbf{Continual-MAML} first pre-trains with MAML, obtanining $\phi$. At continual-learning time, the model adapts $\phi$ to new distributions. The algorithm retrains its slow weights $\phi$ when it detects an OoD task to add new knowledge to the model. (Figure is adapted from Figure 1 in \citet{finn2017model}.)}
    \label{fig:cartoon}
\end{figure*}

We propose Continual-MAML (see \cref{fig:cartoon}), a CL baseline based on MAML~\cite{finn2017model} that can cope with the challenges of \newacron{}. Continual-MAML (see \cref{algo:cmaml} or its complete version \cref{algo:cmaml_full} in \cref{app:algo}) consists of two stages: pre-training and continual learning.

The pre-training phase consists of MAML. That is, meta-learning model parameters such that a small number of gradient steps on a small new task will produce good generalization performance on that task (\cref{algo:cmaml_full}, L6--13). Specifically, the model adapts its initial weights $\phi$ to multiple tasks in the inner loop, obtaining $\theta$. Then it updates the initialization $\phi$ in the outer loop. Note that the inner loop learning rate is meta-learned ($\phi_\eta$ in \cref{algo:cmaml_full}, L10).

At CL time (\cref{algo:cmaml}), the inner loop optimization adapts the model to the current task. Specifically, the model uses current data $X_t,Y_t$ to obtain fast weights $\theta_t$ (\cref{algo:cmaml}, L21). Assuming that the data is locally stationary, it makes a prediction on the following data $X_{t+1}$ and incurs a loss (\cref{algo:cmaml}, L18). In the case of a sudden distribution shift, the model will fail at its first prediction because its fast weights $\theta_t$ are not suited for the new task yet, but it will have recovered by the next. The recovery is achieved by learning new fast weights $\theta_{t+1}$ once the algorithm gets feedback on its prediction (\cref{algo:cmaml}, L25). Note that for some real-life applications, this feedback could be delayed~\cite{kaelbling1996reinforcement}. Finally, to accumulate new knowledge, we further update the meta parameters $\phi$ on the incoming data as well (\cref{algo:cmaml}, L24).

We also propose two features to improve Continual-MAML's performance. First, the algorithm must update its knowledge only when it is solving an OoD task. Accordingly, we introduce a hyperparameter $\lambda$ that controls the behavior of the algorithm between never training on the incoming data at CL time to always training (MAML and C-MAML in \Cref{sec:experiments}). Specifically, when $\mathcal{L}(f_{\theta_{t-1}}(X_t), Y_t)\!>\!\lambda$, new knowledge is incorporated through outer loop optimization of the learned initialization. This mechanism is exemplified in \Cref{fig:cartoon}. To obtain a smoother interpolation between behaviors, we opted for a soft relaxation of the mechanism (\cref{algo:cmaml}, L23) where $g_\lambda: \mathbb{R} \rightarrow (0,1)$.  We call this first feature \textit{\firstFeature{}} (\firstFeatureAcron{}). 

Second, to further leverage the local stationarity of OSAKA, we introduced a mechanism that keeps fine-tuning the fast weights $\theta$ (\cref{algo:cmaml}, L21) until a context shift or task boundary is detected. The simple yet effective context shift detection mechanism works by monitoring the difference in loss with respect to the previous task and is controlled by a hyperparameter $\gamma$ (\cref{algo:cmaml}, L20). We call this second feature \textit{\secondFeature{}} (\secondFeatureAcron{}). In practice, we use a buffer to accumulate data whilst no task boundary is detected such that we can update the slow weights $\phi$ with more examples once it's detected (see \cref{algo:cmaml_full} in \cref{app:algo}). One can think of the update after the task boundary detection as a knowledge consolidation phase.

An ablation of both mechanisms and an hyperparameter sensitivity analysis are provided in \Cref{ssec:results} and \Cref{app:hparam_analysis}, respectively. 

As a result, different from previous CL literature, the proposed algorithm benefits from fast adaptation, dynamic representations, task boundary detection, and computational efficiency, as we describe next.

\paragraph{Fast Adaption} During pre-training, Continual-MAML learns a weight initialization that adapts fast to new tasks. This is different from CL methods that focus on incorporating as much knowledge as possible into one representation that has to maximize performance in a multi-task regime.

\paragraph{Dynamic representations} In \newacron{}, significant distribution shifts occur periodically. As shown in \Cref{sec:experiments}, models that require a fixed representation would fail to adapt. Instead, Continual-MAML, equipped with \firstFeatureAcron{}, detects OoD data and then learns new knowledge using outer-loop optimization.

\paragraph{Computational efficiency} As described by \citet{Farquhar18}, CL agents should operate under restricted computational resources since remembering becomes trivial in the infinite-resource setting. Continual-MAML satisfies this desideratum by allowing the agent to forget (to some extent) and re-allocate parametric capacity to new tasks. Likewise, no computationally expensive mechanisms, such as replay \cite{Chaudhry19}, or BGD \cite{zeno2018task,He2019TaskAC}, are used to alleviate catastrophic forgetting in our method.

\paragraph{Task boundary detection} Continual-MAML detects context shifts which not only help to condition its predictive function on more datapoints (\secondFeatureAcron{}), it also avoids mixing gradient information from two different distributions.

%------------------------------------------- RELATED ------------------------------------------------------%
%----------------------------------------------------------------------------------------------------------%
% \vspace{-0.2cm}
\section{Related Work}
\label{sec:related_work}
% \vspace{-0.2cm}

Continual Learning (CL) \citep{mccloskey1989catastrophic,thrun1995lifelong} has evolved towards increasingly challenging and more realistic evaluation protocols. The first evaluation frameworks \citep{Goodfellow13,kirkpatrick2017overcoming} were made more general in \cite{zeno2018task,Aljundi2019TaskFreeCL} via the removal of the known task boundaries assumption. Later, \cite{He2019TaskAC} proposed to move the focus towards \textit{faster remembering}, or continual-meta learning (CML), which measures how quickly models recover performance rather than measuring the models' performance without any adaptation. \newacron{} builds upon this framework to get closer to real-life applications of CL, as explained in \Cref{sec:osaka}. 

\citet{Harrison2019ContinuousMW} propose a new CML framework and accompanying model (MOCA). OSAKA shares commonalities with this framework, but they are fundamentally different: it does not (1) allow context-dependent targets, (2) expose the algorithms to OoD tasks at CL time, (3) allow new unknown labels, nor (4) propose an update CL evaluation protocol.
Further details are in \Cref{app:moca}.

% --------------------------------------------- EXPERIMENTS -----------------------------------------------------%
%----------------------------------------------------------------------------------------------------------------%

\section{Experiments}
\label{sec:experiments}

We study the performance of different baselines in the proposed \newacron{} setup. We first introduce the datasets, methods, and baselines, and then report and discuss experimental results and observations.

\subsection{Experimental setup}
\label{ssec:datasets}

For all datasets we study two different levels of non-stationarity at CL time, $\alpha$ values of 0.98 and 0.90. Unless otherwise stated the continual-learning episodes have a length of 10,000 timesteps; the probability to visit the pre-training distribution and to visit one of the OoD ones is 0.5 and 0.25, respectively; we report the performance averaged over 20 runs per model and their standard deviation. Statistical significance is assessed using a 95\% confidence interval and highlighted in bold. Further experimental details are provided in \Cref{app:exp_details}. We now introduce our three datasets. A few examples from each are shown in \Cref{app:datasets_baselines}. % \Cref{fig:datasets}.

\paragraph{Omniglot / MNIST / FashionMNIST}
In this study, we pre-train models on the first 1,000 classes of Omniglot~\cite{lake2015human}. At CL time, the models are exposed to the full Onniglot dataset, and two out-of-distribution datasets: MNIST~\cite{lecun-mnisthandwrittendigit-2010} and FashionMNIST~\cite{xiao2017fashion}. Concerning the reported performance, MNIST is a simpler dataset than Omniglot, and FashionMNIST is the hardest. During CL time, the tasks switch with probability $1 \!- \!\alpha$. For this study, we sample 10-way 1-shot classification tasks.

\vspace{0.4cm}

\paragraph{Synbols} In this study, models are pre-trained to classify characters from different alphabets on randomized backgrounds~\cite{lacoste2020synbols}. Tasks consist of 4 different symbols with 4 examples per symbol. During CL time, the model is exposed to a new alphabet. Further, the model will have to solve the OoD task of font classification, where the input distribution does not change, only its mapping to the output space. The font classification task consists of 4 different fonts with 4 symbols per font.

\paragraph{Tiered-ImageNet} Like Omniglot, Tiered-ImageNet~\cite{ren2018meta}  groups classes into super-categories corresponding to higher-level nodes in the ImageNet \cite{deng2009imagenet} hierarchy (we use 20/6/8 disjoint sets for training/validation/testing nodes). We use these higher-level splits to simulate a shift of distribution. We follow the original splits, where the test set contains data that is out of the training and validation distributions. Thus, we use their training set for pre-training, and introduce their validation and test sets at CL time. We refer to them as train, test and OoD in \Cref{tab:tiered}, respectively. Since only one of the two introduced sets is OoD, we increase its probability of being sampled to 0.5, in accordance to the previous benchmarks. This experiment uses 20,000 steps (twice as the others).

\subsection{Baselines}
\label{ssec:baselines}

%\Cref{tab:baselines}
\Cref{app:datasets_baselines} compares the main features of the baselines we benchmark in the \newacron{} setting. For meta-learning methods, ADAM \cite{kingma2014adam} and SGD are used for the outer and inner updates, respectively. 

\paragraph{Online ADAM and Fine tuning.} We use ADAM without and with pre-training as a lower bounds.

\paragraph{BGD \citep{zeno2018task}.} Bayesian Gradient Descent (BGD) is a continual learning algorithm that models the distribution of the parameter vector $\phi$ with a factorized Gaussian. Similarly to \cite{He2019TaskAC} we apply BGD during the continual learning phase. More details about this baseline are provided in \Cref{app:bgd}.

\paragraph{MAML~\citep{finn2017model}.} MAML consists of a pre-training stage and a fine-tuning stage. During pre-training, the model learns a general representation that is common between the tasks. In the fine-tuning stage, the model fine-tunes its layers to adapt to a new task.

\paragraph{ANIL~\cite{raghu2019rapid}.} ANIL differs from MAML only in the fine-tuning stage. Instead of adapting all the network layers, ANIL adapts only the network's head towards the new task. The goal of this baseline is to show the problem with static representations in the continual learning setup. Therefore, ANIL is representative of meta-continual learning.

\paragraph{MetaBGD and MetaCOG~\cite{He2019TaskAC}.} MetaBGD performs CML using MAML and BGD to alleviate catastrophic forgetting. MetaCOG introduces a per-parameter mask learned in the inner loop.

\subsection{Experimental results}
\label{ssec:results}

\begin{figure}
  \begin{subfigure}[b]{0.33\textwidth}
    \includegraphics[trim={0.cm 0.cm 0.cm 0.cm}, width=\textwidth]{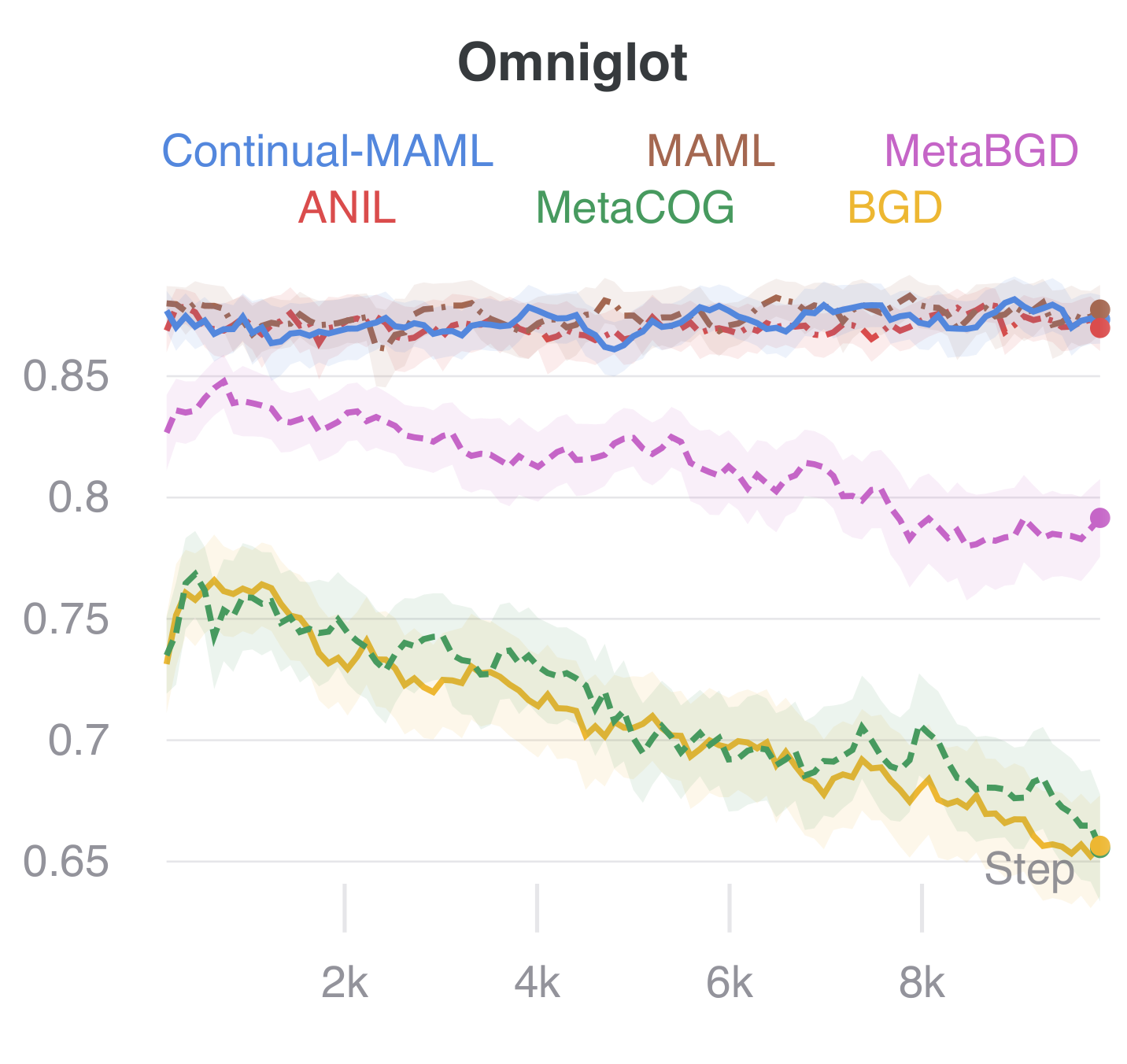}
  \end{subfigure}
  \begin{subfigure}[b]{0.33\textwidth}
    \includegraphics[trim={0.cm 0.cm 0.cm 0.cm}, width=\textwidth]{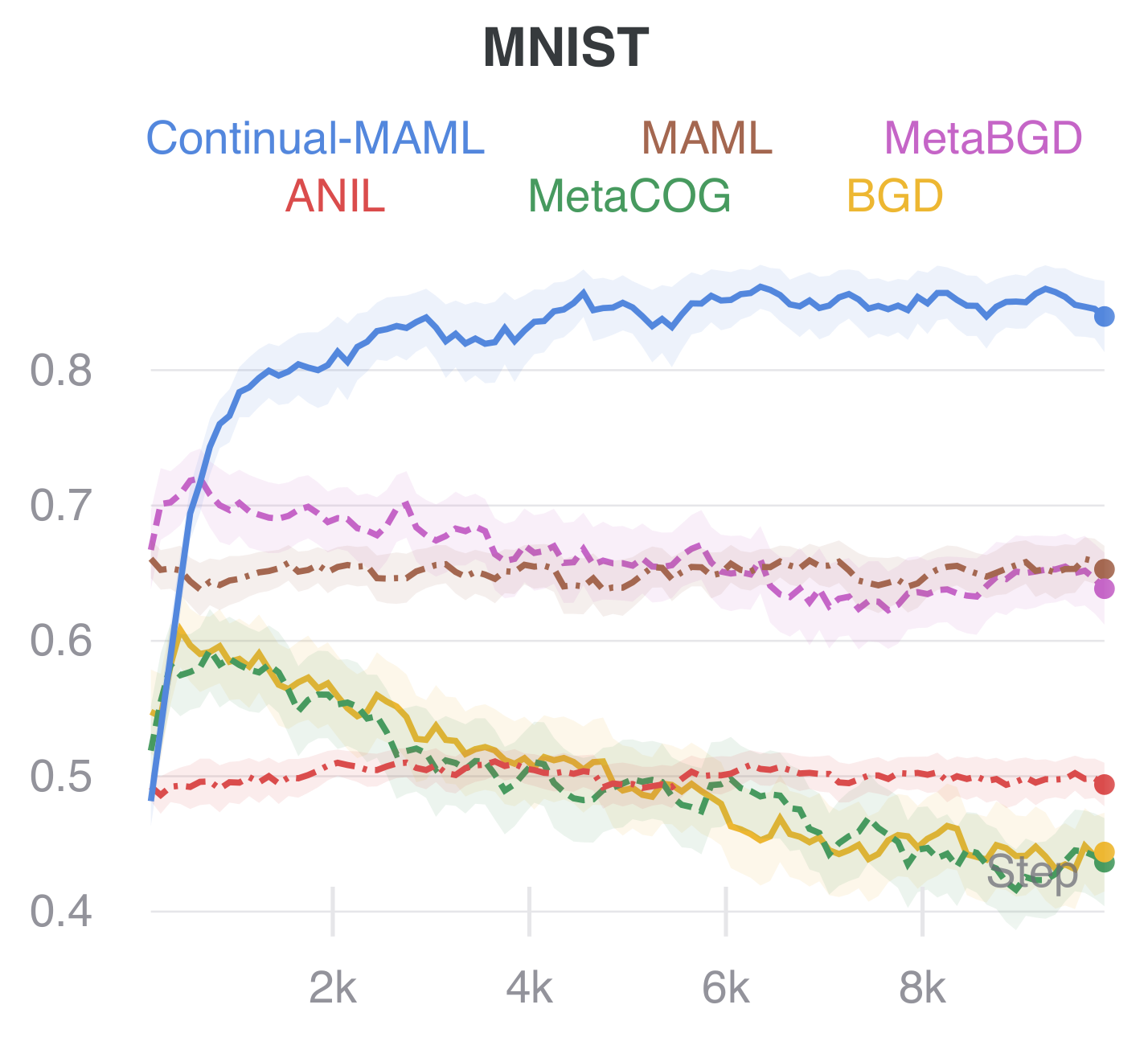}
  \end{subfigure}
  \begin{subfigure}[b]{0.33\textwidth}
    \includegraphics[trim={0.cm 0.cm 0.cm 0.cm}, width=\textwidth]{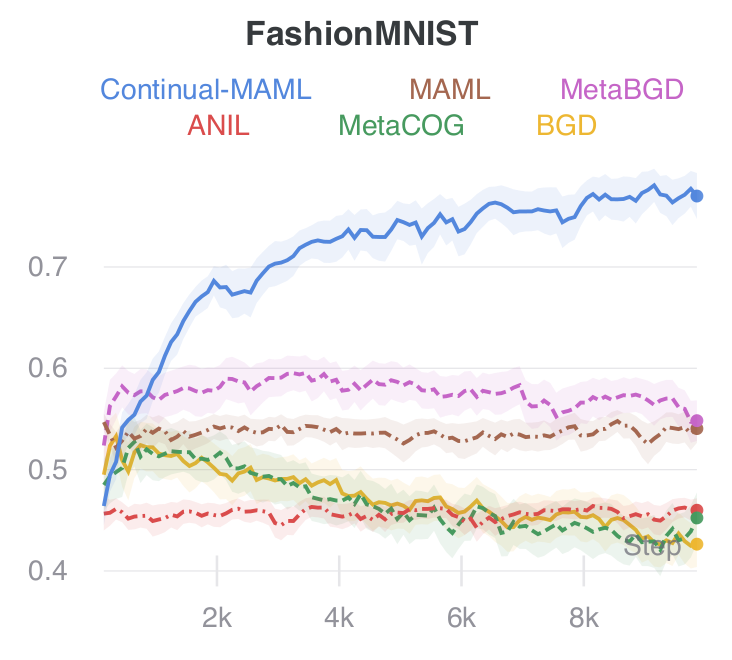}
  \end{subfigure}
  \caption{\textbf{Omniglot / MNIST / FashionMNIST experiment} in the $\alpha\!=\!0.90$ regime. Methods are allowed pre-training on Omniglot before deployment on a stream of Omniglot, MNIST and FashionMNIST tasks. We report the online performance (not cumulative) at each time-step with averaged over 20 runs, as well as standard error. Online ADAM and Fine tuning lie below of the graph. Continual-MAML is the only method with enough plasticity to increase its performance on new tasks, i.e. from MNIST and FashionMNIST, whilst simultaneously being stable enough remember the pretraining tasks, i.e. from Omniglot.}
  \label{fig:omniglot}
\end{figure}

%%%% RESULTS on synbols
\setlength{\tabcolsep}{0.85pt}
\begin{table*}[ht]
  %\vskip 0.15in
  %\begin{tiny}
  \centering
  \centerline{
  \begin{scriptsize}
  \begin{sc}
  %\resizebox{\textwidth}{!}{%
    \begin{tabular}{l|cccc|cccc}
    \toprule
    & \multicolumn{4}{c|}{$\alpha=0.98$} & \multicolumn{4}{c}{$\alpha=0.90$} \\
    model & Total & Prev. Alph. & New Alph. & Font Class. & Total & Prev. Alph. & New Alph. & Font Class. \\ 
    \midrule 
    %\hline
        Online ADAM & 59.6  \std{1.5} & 63.7  \std{2.3} & 59.5  \std{3.7} & 50.7  \std{2.9} & 27.5  \std{0.8} & 28.3  \std{1.1} & 26.3  \std{0.9} & 26.9  \std{0.7} \\
        Fine tuning & 64.0  \std{2.0} & 69.6  \std{2.1} & 63.0  \std{3.6} & 52.9  \std{2.8} & 26.6  \std{1.8} & 27.0  \std{2.4} & 26.2  \std{1.5} & 26.1  \std{1.2} \\
        MAML \citep{finn2017model} & 71.2  \std{2.8} & 90.3  \std{0.8} & 65.7  \std{1.5} & 37.9  \std{1.1} & 69.3  \std{0.9} & 86.3  \std{0.5} & 64.3  \std{0.7} & 40.4  \std{0.7} \\
        ANIL \citep{raghu2019rapid} & 69.4  \std{1.9} & 91.3  \std{0.8} & 59.0  \std{1.6} & 33.2  \std{1.0} & 70.2  \std{0.8} & \bf88.4  \std{0.4} & 68.7  \std{0.6} & 35.1  \std{0.5} \\
        BGD \citep{zeno2018task} & 68.3  \std{1.4} & 73.6  \std{2.3} & 69.7  \std{3.0} & 56.1  \std{3.5} & 33.9  \std{1.3} & 36.7  \std{1.6} & 32.0  \std{1.9} & 30.3  \std{0.9} \\
        % MAML & 71.2  \std{2.8} & 90.3  \std{0.8} & 65.7  \std{1.5} & 37.9  \std{1.1} & 69.3  \std{0.9} & 86.3  \std{0.5} & 64.3  \std{0.7} & 40.4  \std{0.7} \\
        MetaCOG \citep{He2019TaskAC} & 68.3  \std{1.7} & 73.6  \std{1.7} & 69.6  \std{2.8} & 56.8  \std{2.8} & 34.6  \std{1.3} & 37.1  \std{1.8} & 33.5  \std{2.4} & 30.5  \std{1.0} \\
        MetaBGD \citep{He2019TaskAC} & 72.5  \std{1.6} & 77.8  \std{1.8} & 73.6  \std{1.7} & 58.8  \std{3.5} & 60.3  \std{0.4} & 65.8  \std{0.7} & 62.2  \std{1.4} & 47.8  \std{1.4} \\
        % MetaCOG & 68.3  \std{1.7} & 73.6  \std{1.7} & 69.6  \std{2.8} & 56.8  \std{2.8} & 34.6  \std{1.3} & 37.1  \std{1.8} & 33.5  \std{2.4} & 30.5  \std{1.0} \\
        % fine\_tuning & 64.0  \std{2.0} & 69.6  \std{2.1} & 63.0  \std{3.6} & 52.9  \std{2.8} & 26.6  \std{1.8} & 27.0  \std{2.4} & 26.2  \std{1.5} & 26.1  \std{1.2} \\
        % online\_sgd & 59.6  \std{1.5} & 63.7  \std{2.3} & 59.5  \std{3.7} & 50.7  \std{2.9} & 27.5  \std{0.8} & 28.3  \std{1.1} & 26.3  \std{0.9} & 26.9  \std{0.7} \\
        \midrule %\hline
        C-MAML & 74.4  \std{1.4} & 79.4  \std{1.1} & 76.3  \std{2.6} & 61.6  \std{3.1} & 61.2  \std{2.5} & 66.5  \std{3.1} & 62.9  \std{2.8} & 49.3  \std{1.7} \\
        C-MAML + pre. & 78.4  \std{1.0} & 86.6  \std{1.0} & 78.2  \std{1.4} & 60.9  \std{2.6} & 73.3  \std{1.2} & 82.0  \std{1.1} & 75.0  \std{1.6} & 53.8  \std{1.5} \\
        C-MAML + pre. + \firstFeatureAcron{} & 74.8  \std{4.0} & 81.6  \std{6.2} & 75.5  \std{4.5} & 59.5  \std{3.2} & 72.8  \std{0.9} & 81.4  \std{1.2} & 74.4  \std{1.3} & 54.4  \std{1.6} \\
        C-MAML + pre. + \firstFeatureAcron{}+ \secondFeatureAcron{} & \bf 86.3  \std{0.8} & \bf 93.4  \std{0.6} & \bf 86.7  \std{1.8} & \bf 72.0  \std{2.4} & \bf 76.3  \std{0.8} & 84.9  \std{0.7} & \bf 76.4  \std{1.5} & \bf 58.5  \std{1.4} \\
        \bottomrule
    \end{tabular}
  %}
  \end{sc}
  \end{scriptsize}
  }
  %\end{tiny}
  
  \caption{Online cumulative accuracy for the \textbf{Synbols experiments}. Methods are allowed character classification pre-training on an alphabet. Then, they are deployed on a stream of tasks sampled from the pre-training alphabet and a new alphabet, as well as a font classification tasks on the pre-training alphabet. Continual-MAML + pre. outperforms all others methods in total cumulative accuracy and the \secondFeatureAcron{} further increases performance.}

\label{tab:synbols}
\end{table*}

%%%% RESULTS on Tiered
\setlength{\tabcolsep}{2.6pt}
\begin{table*}[ht]
  %\vskip 0.15in
  %\begin{tiny}
  \centering
  \centerline{
  \begin{scriptsize}
  \begin{sc}
  %\resizebox{\textwidth}{!}{% 
    \begin{tabular}{l|cccc|cccc}
    \toprule
    & \multicolumn{4}{c|}{$\alpha=0.98$} & \multicolumn{4}{c}{$\alpha=0.90$} \\
    model & Total & Train & Test & OoD & Total & Train & Test & OoD \\ 
    \midrule
    %\hline
        Online ADAM & 44.5  \std{1.7} & 43.9  \std{2.1} & 44.6  \std{2.2} & 44.6  \std{2.1} & 22.7  \std{0.2} & 22.7  \std{0.4} & 22.6  \std{0.4} & 22.7  \std{0.3} \\
        Fine tuning & 44.6  \std{1.5} & 43.8  \std{2.8} & 44.1  \std{2.1} & 45.2  \std{1.8} & 22.6  \std{0.2} & 22.5  \std{0.3} & 22.7  \std{0.4} & 22.6  \std{0.3} \\
        MAML \citep{finn2017model} & 59.3  \std{1.2} & 61.4  \std{1.9} & 61.0  \std{1.8} & 57.3  \std{1.0} & \bf 60.4  \std{0.4} & \bf 63.2  \std{0.7} & \bf 62.6  \std{0.5} & 58.0  \std{0.3} \\
        ANIL \citep{raghu2019rapid} & 62.4  \std{0.7} & 65.7  \std{0.8} & 64.8  \std{1.3} & 59.5  \std{0.9} & 58.1  \std{0.5} & 61.0  \std{0.8} & 59.7  \std{0.7} & 55.8  \std{0.4} \\
        BGD \citep{zeno2018task} & 54.8  \std{0.8} & 53.8  \std{1.0} & 54.6  \std{1.9} & 55.3  \std{1.0} & 27.7  \std{0.7} & 27.4  \std{0.7} & 27.7  \std{0.8} & 27.8  \std{0.8} \\
        MetaCOG \citep{He2019TaskAC} & 55.2  \std{0.7} & 54.1  \std{1.1} & 55.8  \std{1.6} & 55.4  \std{1.0} & 24.5  \std{0.2} & 23.9  \std{0.4} & 24.0  \std{0.3} & 25.1  \std{0.3} \\
        % MetaBGD \citep{He2019TaskAC} & 42.7  \std{nan} & 42.1  \std{nan} & 43.7  \std{nan} & 42.6  \std{nan} & 46.8  \std{0.8} & 45.8  \std{1.1} & 46.8  \std{1.0} & 47.3  \std{0.9} \\
        MetaBGD \citep{He2019TaskAC} & 55.9  \std{0.6} & 55.7  \std{0.9} & 54.1  \std{1.4} & 56.8  \std{0.9} & 46.8  \std{0.8} & 45.8  \std{1.1} & 46.8  \std{1.0} & 47.3  \std{0.9} \\
        \hline 
        C-MAML & 61.4  \std{0.5} & 59.5  \std{1.4} & 61.2  \std{1.3} & 62.4  \std{0.9} & 53.7  \std{0.3} & 52.0  \std{0.6} & 53.0  \std{0.6} & 54.9  \std{0.5} \\
        C-MAML + pre. & 59.1  \std{0.9} & 57.4  \std{1.2} & 58.4  \std{1.8} & 60.1  \std{1.2} & 57.8  \std{0.7} & 56.3  \std{0.7} & 57.7  \std{0.9} & 58.6  \std{0.7} \\
        C-MAML + pre. + \firstFeatureAcron{} & 66.7  \std{0.9} & 65.7  \std{1.7} & 66.2  \std{1.6} & 67.4  \std{0.9} & 59.7  \std{0.3} & 59.1  \std{0.8} & 59.7  \std{0.6} & \bf 59.9  \std{0.4} \\
        C-MAML + pre. + \firstFeatureAcron{} + \secondFeatureAcron{} & \bf 69.1  \std{0.7} & \bf 68.7  \std{0.9} & \bf 69.3  \std{1.0} &\bf 69.1  \std{1.2} & 53.4  \std{6.4} & 53.5  \std{6.1} & 53.7  \std{6.2} & 53.2  \std{6.6} \\
        \bottomrule
    \end{tabular}
  %}
  \end{sc}
  \end{scriptsize}
  }
  %\end{tiny}
  \caption{Online cumulative accuracy for the \textbf{Tiered Imagenet experiment} (see \cref{ssec:datasets} for the experimental details). For this experiment, Continual-MAML outperforms others methods in the more non-stationary regime ($\alpha\!=\!0.98$). However, in the less-nonstationary one, MAML achieves better results due to its higher stability. Additionally, the \firstFeatureAcron{} mechanism consistently improved Continual-MAML's performance. }
\label{tab:tiered}
\end{table*}

For all benchmarks, we report results on two $\alpha$-locally-stationary environments. The first benchmark's results show online accuracy as function of timesteps in \Cref{fig:omniglot} (full results are found in \Cref{app:omni}). For Synbols and Tiered-Imagenet, the average accuracies over time are reported in \Cref{tab:synbols,tab:tiered}, respectively. For both regimes, the first column is the average performance over all predictions. The second, third and fourth columns show the performance on the three different settings. The prefix {\small{PRE.}} stands for pretraining. Algorithms perform better in the more locally-stationary regime ($\alpha\!=0\!.98$) because they spend more time in each task before switching.  

\paragraph{Fast adaptation} We found fast adaptation (or meta-learning) to be the most critical feature for models to perform well in OSAKA, as highlighted by the performance gap between Online ADAM and Continual-MAML (up to +33\% in Synbols $\alpha\!=\!0.90$). This gain comes from two advantages: quickly changing weights after a task/context switch, having slow ($\phi$), and fast ($\theta$) weights, which alleviate catastrophic forgetting.

\paragraph{Dynamic representations} Next, models need the ability to adapt the embedding space to correctly classify the OoD data. The Synbols font classification task highlights that learning a new mapping from the same inputs to a new output space is challenging when the embedded space is static. Namely, the dynamic representations of Continual-MAML offer a $23.7\%$ and a $28.4\%$ improvement in $\alpha \! = \! 0.98$ compared to MAML and ANIL. This behavior is demonstrated in \Cref{fig:omniglot} were these two baselines do not improve their performances over time, which is precisely the goal of CL. Thus, these results demonstrates the inapplicability of current MCL to real scenarios. Although MCL can continually learn new tasks without forgetting, its static embedded space will prevent it from learning tasks lying outside of the pre-training data distribution.

\paragraph{Computational efficiency} Moreover, adding BGD to slow-down forgetting hinder the acquisition of new knowledge. Removing this feature, e.g. from MetaBGD to Continual-MAML, increases the performance in five out of six experiments and diminishes the computation cost by 80\%. 

\paragraph{\FirstFeature{}} We now analyse, via ablations, the mechanisms we added to Continual-MAML for further improvements. Modulating the updates improved the performance in Omniglot and Tiered experiments but decreased it in Synbols' (C-MAML {\small{+ PRE.}} vs. C-MAML + {\small{PRE. + UM}}, resulting in an average increase of 1.7\%. In Appendix \ref{app:hparam_analysis}, we show how this mechanism interpolates C-MAML {\small{+ UM}}'s behavior betweeen MAML and C-MAML.  

\paragraph{\SecondFeature{}} Finally, our \secondFeatureAcron{} enabled by the task boundary detection mechanism helps achieve impressive gains in the locally more stationary regime (+11.5\% and 2.4\% in Synbols and Tiered-ImageNet, respectively). In the other regime ($\alpha=0.90$), the shorter task sequences limits the room for improvements and the results are inconclusive. 
An hyperparameter sensitivity analysis on $\gamma$ (see \cref{app:hparam_analysis}) in terms of precision and recall for boundary detection accuracy shows that difference in loss magnitudes (see \cref{algo:cmaml} L20) is a good signal for detecting context shifts.

%--------------------------------------------- CONCLUSION --------------------------------------------------%
%-----------------------------------------------------------------------------------------------------------%

\section{Conclusions}
\label{sec:conclusions}

We propose \newacron~a new approach to continual learning that focuses on online adaptation, faster remembering and is aligned to real-life applications. This framework is task agnostic, allows context-conditioned targets and task revisiting. Furthermore, it allows pre-training, and introduces OoD tasks at continual-learning time. 
We show that the proposed setting is challenging for current methods that were not designed for \newacron{}. We introduce Continual-MAML, an initial baseline that addresses the challenges of \newacron{} and we empirically demonstrate its effectiveness.

% NOTE: make it longer for Arxiv

% \newpage
%-------------------------------------------- BROADER ------------------------------------------------------%
%-----------------------------------------------------------------------------------------------------------%

\section*{Broader Impact}

Our work proposes a more-realistic (synthetic) continual-learning environment. This research could help accelerate the deployment of CL algorithms into applications such as autonomous driving, recommendation systems, information extraction, anomaly detection, and others. A domain often associated with continual learning is health care. 
In health care, patient data is (usually) very sensitive, CL algorithms can be the solution to accumulating knowledge from different hospitals: they can be trained continually across hospitals without the data ever leaving the premise.

\textbf{Possible negative impacts:} 
Our framework enables previous tasks to be forgotten at different rates. If data is patient-level data and different tasks relate to different subset of patients, then it means that the system's performance on past patients could vary. A diagnostic system, for example, could forget how to properly diagnose a patient from a previous population whilst learning about a new one. Further research, possibly at the intersection of continual learning and fairness, is needed before the safe deployment of these algorithms.

\textbf{Possible positive impacts:} 
The aforementioned negative impact may also be its greatest asset for having a positive impact. 
Returning to our example, practitioners could understand to what extent a diagnostic system  forgets previous diagnostics. They could then use and develop OSAKA to calibrate their algorithms to match their desiderata (e.g., by choosing when the negative consequence of forgetting may outweight the benefits of additional training data).

\begin{ack}

Laurent Charlin is supported through a CIFAR AI Chair and grants from NSERC, CIFAR, IVADO, Samsung, and Google. Massimo Caccia is also supported through a MITACS grant. We would like to thank Grace Abuhamad for an helpful discussion on broader impacts.

\end{ack}

%-------------------------------------------- BILBIO -------------------------------------------------------%
%-----------------------------------------------------------------------------------------------------------%
\bibliographystyle{apalike}
\bibliography{ref}

\newpage
%-------------------------------------------- APPENDIX ------------------------------------------------------%
%-----------------------------------------------------------------------------------------------------------%
\input{supplementary}

\end{document}

%% file: supplementary.tex
\appendix

%------------------------------------------- UNIFY ------------------------------------------------------%
%--------------------------------------------------------------------------------------------------------%
\section{A unifying framework}
\label{app:unify}

\begin{table*}[ht]
    \begin{scriptsize}
    % \begin{tiny}
    %\begin{sc}
    \begin{center}
    \resizebox{0.9\linewidth}{!}{%
    \centerline{
    \setlength{\tabcolsep}{6pt}
    \renewcommand{\arraystretch}{1.7}
    \begin{tabular}{ccccc}
    \toprule
        % \hline
        & {\scriptsize Data Distribution} & {\scriptsize Model for Fast Weights} & {\scriptsize Slow Weights Updates} & {\scriptsize Evaluation}  \\
        \midrule
        { \scriptsize Supervised Learning} & $ S, Q \sim C $ & $ f_\theta = \algo(S) $ & --- & $ \loss(f_\theta, Q) $ \\
        \midrule 
        \makecell{\scriptsize Meta-learning} & \makecell{ $\{C_i\}_{i=1}^M \sim \world^M$ \\ $S_i, Q_i \sim C_i $ }& $ f_{\theta_i} = \algo_\phi(S_i)$ & $ \makecell{\nabla_\phi \loss( f_{\theta_i}, Q_i) \\ {\scriptstyle \forall i<N}} $ & $ \sum_{i=N}^M \loss(\algo_\phi(S_i), Q_i) $\\
        %\midrule
        \makecell{\scriptsize  Continual Learning} & $S_{1:T}, Q_{1:T} \sim C_{1:T}$ & $ f_\theta = \cl(S_{1:T}) $ & --- & $\sum_t \loss ( f_\theta, Q_t) $ \\
        \midrule
        \makecell{\scriptsize  Meta-Continual Learning} &  \makecell{ $ \{C_{i,1:T}\}_{i=1}^M \sim \world^M$ \\ $ S_{i, 1:T}, Q_{i, 1:T} \sim C_{i,1:T} $} & $f_{\theta_i} = \cl_\phi(S_{i,1:T}) $ &  \makecell{ $\nabla_\phi \sum_t \loss( f_{\theta_i} , Q_{i,t} )$ \\
         ${\scriptstyle \forall i<N}$ } & $\sum_{i=N}^M\sum_t \loss (\algo_\phi(S_{i,1:T}), Q_{i,t}) $ \\
        %\midrule 
        \makecell{\scriptsize Continual-meta learning}  & $ S_{1:T}, Q_{1:T} \sim C_{1:T}$ & $f_{\theta_t} = \algo_\phi(\highlight{S_{t-1}})$& $\nabla_\phi \loss( f_{\theta_t}, S_t)$ & $ \sum_t \loss( \algo_\phi(S_t), Q_t )$\\ %$ \sum_t \loss( \algo_\phi(S_t), Q_t )$ \\
        \midrule
        \makecell{\scriptsize \newacron{}}  & $ Q_{1:T} \sim C_{1:T}$ & $f_{\theta_t} = \algo_\phi(\highlight{Q_{t-1}})$& $\nabla_\phi \loss( f_{\theta_t}, Q_t)$ &$ \sum_t \loss( f_{\theta_t}, Q_t )$ \\  
        \bottomrule
    \end{tabular}
    %}
    }}
    \end{center}
    % \end{tiny}
    \end{scriptsize}
    \caption{\textbf{A unifying framework} for different machine learning settings. Data sampling, fast weights computation and slow weights updates as well as evaluation protocol are presented with meta-learning terminology, i.e., the support set $S$ and query set $Q$. For readability, we omit OSAKA pre-training.}
    \label{app:uni}
\end{table*}

{\bf Meta-continual learning} combines meta-learning and continual learning. A collection of $M$ sequences of contexts is sampled i.i.d.\ from a distribution over sequences of contexts, $\world^M$, i.e., $ \{ C_{i,1:T}\}_{i=1}^M \sim \world^M$ and $S_{i,1:T}, Q_{i,1:T} \sim
X_{i,1:T}\mid C_{i,1:T}$. Next, the continual learning algorithm, $\cl_\phi$, can be learned using the gradient $\nabla_\phi \sum_t \loss( \cl_\phi( S_{i,1:T} ) , Q_{i,t} )$, for $i < N < M$ and evaluated on the remaining sets $\sum_{i=N}^M \sum_t \loss( \cl_\phi( S_{i,1:T} ) , Q_{i,t} )$. As in continual learning, the target distribution is fixed.

{\bf Continual-meta learning} considers a sequence of datasets $S_{1:T}, Q_{1:T} \sim C_{1:T}$. At training or continual-learning time, $S_{1:T}$ is both used as a support and query set: $S_t$ is used as the query set and $S_{t-1}$ as the support. Predictions at time $t$ are made using $f_{\theta_t} = \algo_\phi(Q_{t-1})$. Since local stationarity is assumed, the model always fails on its first prediction when the task switches. Next, using $l_t = \loss(f_{\theta_t}, S_t)$, the learning of $\phi$ is performed using gradient descent of $\nabla_\phi l_t$. The evaluation is performed at the end of the sequence where $\algo_\phi$ recomputes fast weights using the previous supports and is tested on the query set, i.e., $\sum_t \loss( \algo_\phi(S_t), Q_t)$. Similar to meta-learning, continual-meta learning allows for context-dependent targets.

\clearpage
%------------------------------------------- ALGOS ------------------------------------------------------%
%--------------------------------------------------------------------------------------------------------%
\section{Algorithms}
\label{app:algo}

\scalebox{0.85}{
    \begin{minipage}{0.7\linewidth}
    \setlength{\textfloatsep}{3mm}
    \begin{algorithm2e}[H]
    \SetAlgoLined
    \DontPrintSemicolon
    \textbf{Require:}  $P(C_\text{pre})$, $P(C_\text{cl})$: distributions of contexts (or tasks) \\
    \textbf{Require:} $\gamma$, $\lambda$: threshold and regularization hyperparameters \\
    \textbf{Require:} $\eta$: step size hyperparameter \\
    \textbf{Initialize:} $\phi$, $\theta$: Meta and fast adaptation parameters \\
    \textbf{Initialize:} $\eta_\phi$: learnable inner loop learning rate \\
    \textbf{Initialize:} $\mathcal{B}$: buffer of incoming data \\ 
    \While{ pre-training }{
        % MAML updates as in \cite{finn2017model}, Algorithm 1 \\
        Sample batch of contexts (or tasks) $\{C_i\}_{i=1}^B \sim P(C_\text{pre})$ \\
            \ForEach{ $C_i$}{
                Sample data from context $\bm{x}_i, \bm{y}_i \sim P(\bm{x},\bm{y} | C_i)$ \\
                $\theta_i \gets \phi - \phi_\eta \nabla_\phi \mathcal{L}\big(f_\phi(\bm{x}_i[:k]), \bm{y}_i[:k]\big)$ \\
        }
    $\phi \gets \phi - \eta \nabla_\phi \sum_i \mathcal{L}\big(f_{\theta_i}(\bm{x}_i[k:]), \bm{y}_i[k:]\big)$ \\
    }
    \textbf{Initialize:} current parameters $\theta_0 \gets \phi$ \\
    \While{ continually learning }{
        Sample current context $C_t \sim P(C_\text{cl} | C_{t-1})$ \\
        Sample data from context $\bm{x}_t, \bm{y}_t \sim P(C_t)$ \\
        Incur loss $\mathcal{L}\big(f_{\theta_{t-1}}(\bm{x}_t), \bm{y}_t\big)$ \\
        Virtual model $\tilde{\theta_t} \gets \phi - \phi_\eta \nabla_\phi \mathcal{L}\big(f_\phi(\bm{x}_t), \bm{y}_t\big)$ \\
        \uIf{$\mathcal{L}\big(f_{\theta_{t-1}}(\bm{x}_t), \bm{y}_t\big) - \mathcal{L}\big(f_{\tilde{\theta_{t}}}(\bm{x}_t), \bm{y}_t\big) < \gamma$}{
         \# No context shift detected \\
         Further fine tune the fast parameters $\theta_t \gets \theta_{t-1} - \phi_\eta \nabla_\theta \mathcal{L}\big(f_{\theta_{t-1}}(\bm{x}_t), \bm{y}_t\big)$ \\
         Add $(\bm{x}_t, \bm{y}_t)$ to buffer $\mathcal{B}$ \\
        }
       \uElse{
         \# Task boundary detected \\
        Sample training data from buffer $\bm{x}_{\text{train}}, \bm{y}_{\text{train}} \sim \mathcal{B}$ \\
        Fast adaptation $\theta \gets \phi - \phi_\eta \nabla_\phi \mathcal{L}\big(f_\phi(\bm{x}_{\text{train}}), \bm{y}_{\text{train}}\big)$ \\
        sample test data from buffer $\bm{x}_{\text{test}}, \bm{y}_{\text{test}} \sim \mathcal{B}$ \\
        Modulated learning rate $\eta_t \gets  \eta g_\lambda \Big( \mathcal{L}\big(f_{\theta}(\bm{x}_{\text{test}}), \bm{y}_{\text{test}}\big) \Big)$ \\
        Update Meta parameters $\phi \gets \phi - \eta_t \nabla_\phi \mathcal{L}\big(f_{\theta}(\bm{x}_{\text{test}}), \bm{y}_{\text{test}}\big)$ \\
        Reset buffer $\mathcal{B}$ \\
        Reset fast parameters $\theta_t \gets \phi - \phi_\eta \nabla_\phi \mathcal{L}\big(f_\phi(\bm{x}_t), \bm{y}_t\big)$  \\
        }
        $t \gets t+1$ \\
    }
    \caption{Continual-MAML}
\label{algo:cmaml_full}
\end{algorithm2e}
\end{minipage}
} 
 
\scalebox{0.85}{
    \begin{minipage}{0.7\linewidth} 
    \setlength{\textfloatsep}{3mm}
    \begin{algorithm2e}[H]
    \SetAlgoLined
    \DontPrintSemicolon
    \textbf{Require:}  $P(C_\text{pre})$, $P(C_\text{cl})$: distributions of contexts (or tasks) \\
    \textbf{Require:} $\gamma$, $\lambda$: threshold hyperparameters \\
    \textbf{Require:} $\eta$: step size hyperparameter \\
    \textbf{Initialize:} $\phi$, $\theta$: Meta and fast adaptation parameters \\
    \While{ pre-training }{
        % MAML updates as in \cite{finn2017model}, Algorithm 1 \\
    Sample batch of contexts (or tasks) $\{C_i\}_{i=1}^B \sim P(C_\text{pre})$ \\
        \ForEach{ $C_i$}{
        Sample data from context $\bm{x}_i, \bm{y}_i \sim P(C_i)$ \\
        $\theta_i \gets \phi - \phi_\eta \nabla_\phi \mathcal{L}\big(f_\phi(\bm{x}_i[:k]), \bm{y}_i[:k]\big)$ \\
        }
    $\phi \gets \phi - \eta \nabla_\phi \sum_i \mathcal{L}\big(f_{\theta_i}(\bm{x}_i[k:]), \bm{y}_i[k:]\big)$ \\
    }
    \textbf{Initialize:} current parameters $\theta_0 \gets \phi$ \\
    \While{ continually learning }{
        Sample current context $C_t \sim P(C_\text{cl} | C_{t-1})$ \\
        Sample data from context $\bm{x}_t, \bm{y}_t \sim P(\bm{x}, \bm{y} | C_t)$ \\
        Incur loss $\mathcal{L}\big(f_{\theta_{t-1}}(\bm{x}_t), \bm{y}_t\big)$ \\
        %  {\color{blue} $\theta_i \gets \phi - \phi_\eta \nabla_\phi \mathcal{L}\big(f_\phi(\bm{x}_i), \bm{y}_i\big)$ } \\
        Reset fast parameters $\theta_t \gets \phi - \phi_\eta \nabla_\phi \mathcal{L}\big(f_\phi(\bm{x}_t, \bm{y}_t\big)$  \\
        \uIf{$\mathcal{L}\big(f_{\theta_{t-1}}(\bm{x}_t), \bm{y}_t\big) - \mathcal{L}\big(f_{\theta_{t}}(\bm{x}_t), \bm{y}_t\big) < \gamma$}{
            \# No task boundary detected \\
            Modulated learning rate $\eta_t \gets  \eta g_\lambda \Big( \mathcal{L}\big(f_{\theta_{i-1}}(\bm{x}_t), \bm{y}_t\big) \Big)$ \\
            $\phi \gets \phi - \eta_t \nabla_\phi \mathcal{L}\big(f_{\theta_{t-1}}(\bm{x}_t), \bm{y}_t\big)$ \\
        }
        $t \gets t+1$ \\
    } 
        \caption{Continual-MAML w/o Prolonged Adaptation Phase}
\label{algo:cmaml_wo_pap}
\end{algorithm2e} 
\end{minipage}}

\clearpage
%------------------------------------------- RELATED ------------------------------------------------------%
%----------------------------------------------------------------------------------------------------------%
\section{Related Work}
\label{app:related_work}

Our method intersects the topics of continual learning, meta learning, continual-meta learning, and meta-continual learning. For each of these topics, we describe the related work and current state-of-the-art methods. 
% We highlight their differences in \ref{tab:related_work}.

\paragraph{Continual learning.} Given a non-stationary data stream, standard learning methods such as stochastic gradient descent (SGD) are prone to catastrophic forgetting as the network weights adapted to the most recent task quickly cannot perform the previous ones anymore. Many continual learning approaches have been proposed in recent years, which can be roughly clustered into: (1) replay-based methods, (2) regularization-based methods, and (3) parameter-isolation methods. 
{\em Replay-based} methods store representative samples from the past, either in their original form (e.g., {\em rehearsal methods} \cite{rebuffi2017icarl, isele2018selective, rolnick2019experience, Aljundi2019Online}, {\em constrained optimization} based on those samples \cite{lopez2017gradient}), or  in a compressed form, e.g., via a generative model \cite{Aljundi2019Online,Caccia2019OnlineLC,DGR,lesort2018generative}. However, those methods require additional storage, which may need to keep increasing when the task sequence is longer. {\em Regularization-based} or {\em prior-based} approaches \cite{kirkpatrick2017overcoming,Nguyen17,zeno2018task} prevent significant changes to the parameters that are important for previous tasks. Most prior-based methods rely on task boundaries. However, they fail to prevent forgetting with long task sequences or when the task label is not given at test time \cite{Farquhar18,Lesort2019RegularizationSF}.
The third family, {\em parameter isolation} or {\em dynamic architecture} methods, attempts to prevent forgetting by using different subsets of parameters for fitting different tasks. This is done either by freezing the old network \cite{xu2018, HAT_Serra} or growing new parts of the network \citep{Dyn_expand_net_Lee,schwarz2018progress}. Dynamic architecture methods, however, usually assume that the task label is given a test time, which reduces their applicability in real-life settings. For more details, recent continual learning surveys have been proposed \cite{de2019continual,Mundt2020AWV}.

\paragraph{Meta learning.} 
Learning-to-learn methods are trained to infer an algorithm that adapts to new tasks \cite{schmidhuber1987evolutionary}. Meta learning has become central for few-shot classification \cite{ravi2016optimization,vinyals2016matching,oreshkin2018tadam}. A commonly used meta-learning algorithm is MAML~\cite{finn2017model}, which optimizes the initial parameters of a network such that adapting to a new task requires few gradient steps. ANIL~\cite{raghu2019rapid} is another variation of meta learning that requires only adapting the network's output layer or head to the new tasks.  These algorithms leverage gradient descent to learn a feature representation that is common among various tasks, but they are not suitable when the new tasks have a drastic distribution shift from the existing tasks. 
% Further, these algorithms suffer from catastrophic forgetfulness if trained on non-i.i.d.\ data, similarly to most machine learning algorithms.
Despite the limitations of meta-learning methods, they can be adapted to address the challenges of continual learning, as we will describe below.

\paragraph{Meta-continual learning.} Since non-stationary data distributions breaks the i.i.d\ assumption for SGD, it is natural to consider continual learning as an optimization problem where the learning rule learns with non-stationary data. Therefore, some recent works focus on learning a non-forgetting learning rule with meta learning, i.e., meta-continual learning.

In \citet{Javed2019Meta}, the model is separated into a representation learning network and a prediction learning network. The representation learning network is meta learned so that the prediction learning part can be safely updated with SGD without forgetting. In \citet{vuorio2018meta}, a gradient-based meta-continual learning is proposed. The update is computed from a parametric combination of the gradient of the current and previous task. This parametric combination is trained with a meta objective that prevents forgetting.

These approaches are all limited by the fundamental assumption of meta learning that the distribution of the meta testing set matches that of the meta training set. Thus it is not guaranteed that the meta-learned representation or update rule is free of catastrophic forgetting when OoD data is encountered in the future. Despite that, meta-continual learning is actively researched \citep{riemer2018learning,beaulieu2020learning}.

\paragraph{Continual-meta learning.} Recently, several methods emerged that address the continual-meta learning setup. FTML~\cite{pmlr-v97-finn19a} extends the MAML algorithm to the online learning setting by incorporating the follow the leader (FTL) algorithm~\cite{hannan1957FTL}. FTL provides an $O(\log T)$ regret guarantee and has shown good performance on a variety of datasets. Dirchlet-based meta learning (DBML) ~\cite{NIPS2019_9112} uses a Dirchlet mixture model to infer the task identities sequentially.

More relevant to our work, MetaBGD~\cite{He2019TaskAC} addresses the problem of fast remembering when the task segmentation is unavailable. MOCA~\cite{Harrison2019ContinuousMW} extends meta-learning methods with a differentiable Bayesian change-point detection scheme to identify whether a task has changed. Continual-meta learning is now an active research field \cite{luo2019learning,antoniou2020defining}.

\clearpage
 \subsection{Contrasting OSAKA and MOCA's framework}
 \label{app:moca}

 In this section, we contrast OSAKA with the recently introduced framework showcasing meta-learning via online changepoint analysis (MOCA) \citep{Harrison2019ContinuousMW}. We are incentivized to discuss these differences because both frameworks can appear similar. Specifically, OSAKA and MOCA's framework represent the tasks or contexts as a hidden Markov chain. However, both settings are fundamentally different and the similarities are superficial. We now highlight their core differences.

 \paragraph{Context-dependent targets} 
In most CL scenarios including  in the MOCA's framework, the joint distribution $p_t(x,y)$ changes through time via the input distribution $p_t(x)$. The target distribution $p(y|x)$ however is fixed (i.e., $p_t(y|x)=p(y|x)$). In other words, in standard incremental CL, new labels still appear even though $p_t(y|x)$ is fixed: they appear via $p_t(x)$ moving its probability mass to new classes. 

OSAKA is more general because it allows for drift in the target distribution $p_t(y|x)$ as well. This is achieved through the latent context variable $C$ as detailed in Section \ref{sec:osaka}. In other words, $p_t(y|x) = p(y|x, c_t)$. This is a common scenarios in partially-observable environments \cite{monahan1982state, garnelo2018conditional} or more generally to any case where a prediction depends on the context, e.g. time-series prediction.

 \paragraph{Out-of-distribution tasks} Similar to \citet{Javed2019Meta,beaulieu2020learning}, MOCA's framework allows for pre-training. However, all those frameworks test their models on similar data at CL time, i.e., new classes from the same dataset. They thus make strong assumptions about the data distribution that the CL agent will be exposed to at deployment time. This assumption can limite the real-world applicability of current methods. 
 
In OSAKA, pre-training is also allowed. However, at CL time, the model will be tested on OoD data distribution w.r.t the pre-training one (see Section \ref{sec:osaka}. OSAKA thus helps us analyze robustness of algorithms to data distribution(s) outside of the pre-training one. It is thus more aligned with real-life cases of CL.

 \paragraph{Expanding set of labels}
In MOCA's framework, all classes are known a priori (see Section B2 in \citet{Harrison2019ContinuousMW}). They do not allow for an expanding set of labels over time, which is a central idea in CL \cite{kirkpatrick2017overcoming, lopez2017gradient, rebuffi2017icarl, Farquhar18, Aljundi2019Online, Aljundi2019Gradient, shin2017continual, Javed2019Meta, Chaudhry19}. MOCA's framework is closer to domain-incremental learning \cite{van2019three}, i.e., classes are fixed but new variations can appear within them.

Similarly to standard CL, OSAKA allows for an expending set of labels. Thus, algorithms' capacity to incrementally learn new concepts is studied in OSAKA.

To conclude, the main contribution of \citet{Harrison2019ContinuousMW} is a new \emph{algorithm}: MOCA. In contrast, OSAKA is a new CL \emph{evaluation framework} aiming to push CL beyond its current limits. We acknowledge that changepoint detection is important for continual learning and refer the readers to \cite{Harrison2019ContinuousMW} for a review of the changepoint detection literature.

\clearpage
%-------------------------------------- DATASETS and BASELINES ------------------------------------------------%
%----------------------------------------------------------------------------------------------------------%
\section{Datasets and Baselines}
\label{app:datasets_baselines}

\centerline{
    \begin{minipage}{0.47\linewidth}
    \includegraphics[trim={3cm 3cm 3cm 3cm},clip,width=\textwidth]{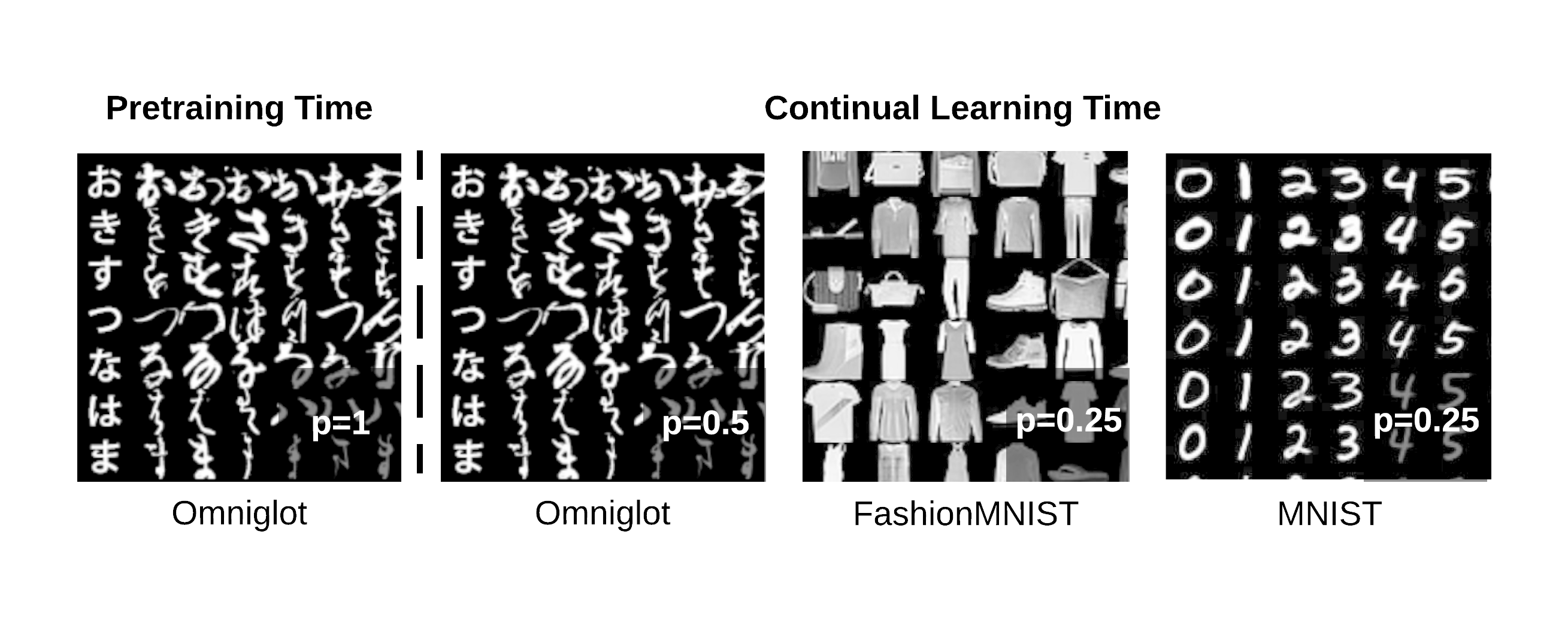}
    \includegraphics[width=\textwidth]{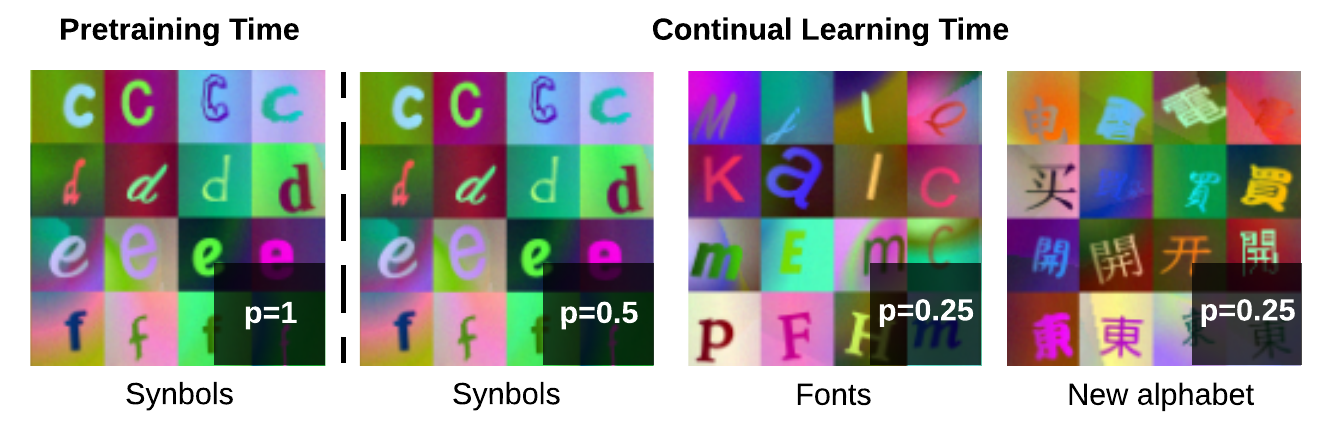}
    \includegraphics[trim={0.25cm 0.25cm 0.25cm 0.5cm},clip,width=\textwidth]{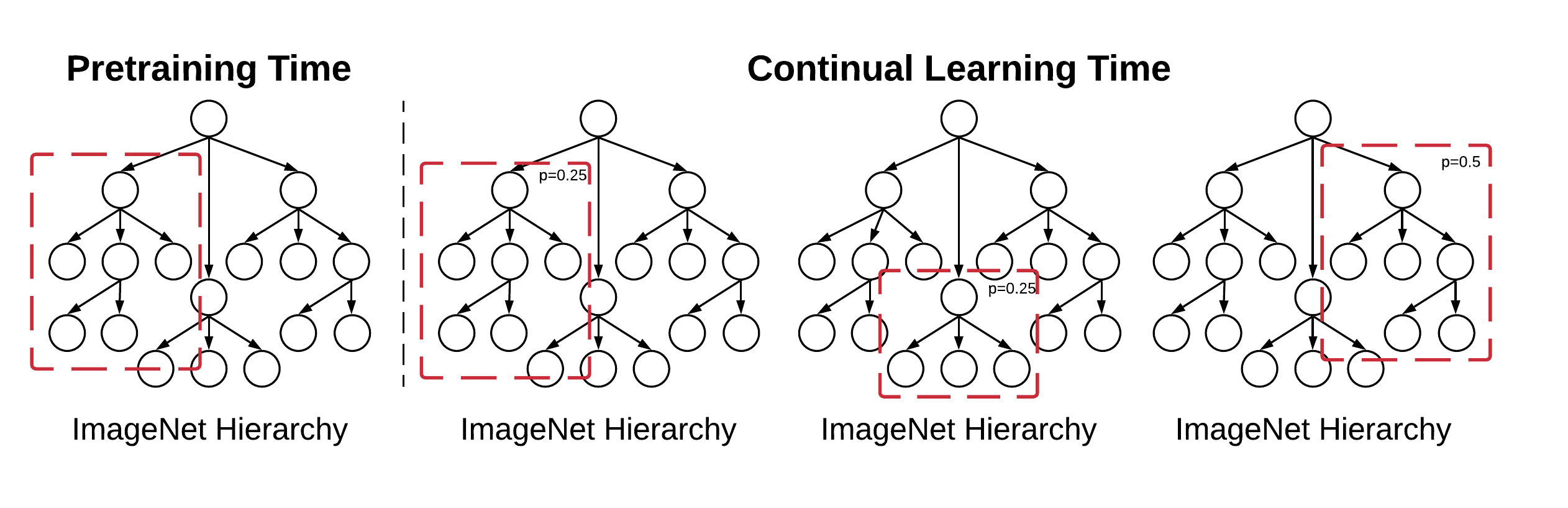}
    \captionof{figure}{\textbf{Benchmarks.} We evaluate our setup on three different benchmarks, each one depicted in one row: Omniglot/MNIST/FashionMNIST, Synbols, and Tiered-ImageNet.}
    \label{fig:datasets}
    \end{minipage}
    \hspace{0.5em}
    \begin{minipage}{0.5\linewidth}
    \centering
    \begin{scriptsize}
    \begin{sc}
        \setlength{\tabcolsep}{1.5pt}
        \setlength{\extrarowheight}{.5em}
        \centering
        \begin{tabular}{lcccccc}
        \toprule
            & \multicolumn{2}{c}{pre-train} & \multicolumn{4}{c}{CL Time} \\
          \cmidrule(l{2pt}r{2pt}){2-3}
          \cmidrule(l{2pt}r{2pt}){4-7}
            model & MAML & ANIL &  MAML  & SGD  & BGD & UM/PAP \\
            \midrule
            Online ADAM                      & $\times$ & $\times$ & $\times$ & $\surd$  & $\times$ & $\times$ \\
            Fine tuning                     & $\times$ & $\surd$  & $\times$ & $\surd$  & $\times$ & $\times$ \\
            BGD \citep{zeno2018task}        & $\times$ & $\times$ & $\times$ & $\times$ & $\surd$  & $\times$ \\
            MAML \citep{finn2017model}      & $\surd$  & $\times$ & $\times$ & $\times$ & $\times$ & n/a  \\
            ANIL \citep{raghu2019rapid}     & $\times$ & $\surd$  & $\times$ & $\times$ & $\times$ & n/a  \\ 
            MetaBGD \citep{He2019TaskAC}    & $\times$ & $\times$ & $\surd$  & $\times$ & $\surd$  & $\times$ \\
            MetaCOG \citep{He2019TaskAC}    & $\times$ & $\times$ & $\surd$  & $\times$ & $\surd$  & $\times$ \\
            Continual-MAML                  & $\surd$  & $\times$ & $\surd$  & $\times$ & $\times$ & $\surd$  \\
            \bottomrule
        \end{tabular}
    \end{sc}
    \end{scriptsize}
    \captionof{table}{\textbf{Baseline comparison.} Columns 2--3 contain pre-training algorithms. Columns 4--7 show training algorithms at continual learning time. UM and PAP stand for \textit{\firstFeature{}} and \textit{\SecondFeature{}}, respectively, and are explained in \Cref{sec:method}.}
    \label{tab:baselines}
    \end{minipage}
}

\vspace{0.5cm}
%-------------------------------------- EXPERIMENT DETAILS ------------------------------------------------%
%----------------------------------------------------------------------------------------------------------%
\section{Experiment Details}
\label{app:exp_details}

The procedure followed to perform the experiments in Section~\ref{sec:experiments} is described next in detail. The code to reproduce the experiments is publicly available at \url{https://github.com/ElementAI/osaka}.

For all experiments, we used a 4-layer convolutional neural network with 64 hidden units as commonly used in the few-shot literature \citep{vinyals2016matching, snell2017prototypical, sung2018learning, rodriguez2020embedding}. All the methods were implemented using the PyTorch library \cite{paszke2019pytorch}, run on a single 12GB GPU and 4 CPUs .

\subsection{Hyperparameter search}
\label{ssec:hparam_search}

Hyperparameters were found by random search. During hyperaparmeter search, we allocated the same amount of trials for each method, i.e., each line in the reported Tables. We used Adam \cite{kingma2014adam} for the outer-loop optimization and SGD in the inner (for meta-learning methods). For each trial, we sampled uniformly a method and then sampled hyperparameters uniformly according to the search space defined in Table \ref{app:hparam_values}. Each for each hyperparameter trial, we ran two continual learning episodes with different seeds. The seeding impacts the neural net initialization as well as what data stream the algorithm will be exposed to. Whenever the first ran didn't return a cumulative accuracy better than random, we omitted the second run. We allocated equal amount of trials to both non-stationary levels $\alpha \in \{0.90, 0.98\}$. We dedicated a fix amount of compute for each benchmarks and further provide specific details in the rest of this section.

\paragraph{Omniglot / MNIST / FashionMNIST} For this benchmark, we allocated a total of 12.5 days of compute. This allowed for 935 trials of which 381 were better than random. 

\paragraph{Synbols} For this benchmark, we allocated a total of 19.5 days of compute. This allowed for 1,309 trials of which 340 were better than random. 

\paragraph{Tiered-Imagenet} For this benchmark, we allocated a total of 62 days of compute. We only ran 1 seed per trials which allowed for 934 trials.

For all benchmarks, concerning the runtime per trials, because BGD requires 5 times more compute than SGD, the BGD baseline took approximately five time longer to run than Online ADAM. Similarly, MetaBGD took approximately 5 time longer to run than C-MAML. Moreover, methods with meta-learning took approximately 5 times longer than methods without.  

We add the following clarification: we do not need a validation set in OSAKA, as there is no \textit{training error}. Specifically, in the CL episodes, algorithms always make prediction on held-out data.

As for the evaluation runs, the best sets of hyperparameters are used to evaluate the methods on 20 new runs. The algorithms are thus exposed to 20 new CL episodes. For clarification, we do not use the best models found in the hyperparameter-search: we only use the hyperparameters to train and evaluate new models.

\begin{table}[h]
% \centerline{
    \centering
    \begin{scriptsize}
    \begin{sc}
        \setlength{\tabcolsep}{3pt}
        \centering
        \begin{tabular}{lcccccccccc}
        \toprule
            model                             & $\eta$  & batch size  & inner-step size & inner iters & first order & MC samples & $\beta$ & $\sigma$ & $\gamma$ & $\lambda$   \\
            \midrule
            Online ADAM                       & $\surd$  & $\times$ & $\times$ & $\times$  & $\times$ & $\times$ & $\times$ & $\times$ & $\times$  & $\times$ \\ 
            Fine Tuning                       & $\surd$  & $\surd$  & $\times$ & $\times$  & $\times$ & $\times$ & $\times$ & $\times$ & $\times$  & $\times$ \\ 
            MAML \citep{finn2017model}        & $\surd$  & $\surd$  & $\surd$  & $\surd$   & $\surd$  & $\times$ & $\times$ & $\times$ & $\times$  & $\times$ \\ 
            ANIL \citep{raghu2019rapid}       & $\surd$  & $\surd$  & $\surd$  & $\surd$   & $\surd$  & $\times$ & $\times$ & $\times$ & $\times$  & $\times$ \\ 
            BGD \citep{zeno2018task}          & $\surd$  & $\times$ & $\times$ & $\times$  & $\times$ & $\surd$  & $\surd$  & $\surd$  & $\times$  & $\times$ \\ 
            MetaBGD \citep{He2019TaskAC}      & $\surd$  & $\surd$  & $\surd$  & $\surd$   & $\surd$  & $\surd$  & $\surd$  & $\surd$  & $\times$  & $\times$ \\ 
            MetaCOG \citep{He2019TaskAC}      & $\surd$  & $\surd$  & $\surd$  & $\surd$   & $\surd$  & $\surd$  & $\surd$  & $\surd$  & $\times$  & $\times$ \\
            Continual-MAML                    & $\surd$  & $\times$ & $\surd$  & $\surd$   & $\surd$ & $\times$  & $\times$ & $\times$ & $\times$  & $\times$ \\
            Continual-MAML + Pre.             & $\surd$  & $\surd$  & $\surd$  & $\surd$   & $\surd$ & $\times$  & $\times$ & $\times$ & $\times$  & $\times$ \\
            Continual-MAML + UM               & $\surd$  & $\times$ & $\surd$  & $\surd$   & $\surd$ & $\times$  & $\times$ & $\times$ & $\surd$   & $\times$ \\         
            Continual-MAML + PAP              & $\surd$  & $\times$ & $\surd$  & $\surd$   & $\surd$ & $\times$  & $\times$ & $\times$ & $\times$  & $\surd$  \\       
            \bottomrule
        \end{tabular}
    \end{sc}
    \end{scriptsize}
    \vspace{0.3cm}
    \caption{\textbf{Method's hyperparameters.} $\eta$ is the step-size or outer-step size for meta-learning methods. Batch size is only needed for methods with pre-training. For methods using meta-learning, we searched the inner-step size, the number of inner iterations (inner iters) and the use of the first order approximation of MAML.  BGD related hyperparameters, i.e., MC samples, $\beta$ and $\sigma$ are explained in Appendix \ref{app:bgd}. $\gamma$ and $\lambda$ are specific of Continual-MAML and operate the update modulation and prolonged adaptation phase mechanisms, respectively. For readability, we omitted 2 hyperparameters related to MetaCOG and refer to the codebase for completeness.}
    \label{tab:baselines_hparams}
% } 
\end{table}

\begin{table}[h]
\centering
\begin{tabular}{@{}l || c c c c c c c c c c@{}}
\toprule 
$\eta$          & 0.0001 & 0.0005 & 0.001 & 0.005 & 0.01 \\
Batch size      & 1 & 2 & 4 & 8 & 16 \\
Inner-step size & 0.0005 & 0.001 & 0.005 & 0.01 & 0.05 & 0.1 & 0.5  \\
Inner iters     & 1 & 2 & 4 & 8 & 16 \\
First Order     & True & False \\
MC Samples      & 5 \\
$\beta$         & 0.5   & 1.0  & 10. \\
$\sigma$        & 0.001 & 0.01 & 0.1 \\
$\gamma$        & 0.25 & 0.5 & 1.0 & 2.0 & 3.0 & 5.0  \\
$\lambda$       & 0.25 & 0.5 & 0.75 & 1.0 & 1.25 & 1.5 & 2.0 & 2.5 & 3.0 \\
\bottomrule
\end{tabular}
\vspace{0.3cm}
\caption{\textbf{Hyperparameter search space.} }
\label{app:hparam_values}
\end{table}

\clearpage
%------------------------------------------- RESULTS ------------------------------------------------------%
%----------------------------------------------------------------------------------------------------------%
\section{Extra Results}
\label{app:more_results}

In this section, we provided further results as well as more details about baselines.

\subsection{Omniglot / MNIST / FashionMNIST}
\label{app:omni}

In Table \ref{tab:omniglot}, we report the full results for the Omniglot / MNIST / FashionMNIST experiment. Contrary to the other experiments, we found that C-MAML pre-training didn't improve results. We thus focus the ablation on C-MAML instead of C-MAML + Pre.

%%%% RESULTS on Omniglot

\setlength{\tabcolsep}{3.5pt}
\begin{table*}[h!]
%   \centering
  \centerline{
  \begin{scriptsize}
  \begin{sc}
    \begin{tabular}{l||cccc|cccc}
    & \multicolumn{4}{c|}{$\alpha=0.98$} & \multicolumn{4}{c}{$\alpha=0.90$} \\
    model & Total & Omniglot & MNIST & Fashion & Total & Omniglot & MNIST & Fashion  \\ 
    \hline
    \hline
        Online ADAM & 73.9  \std{2.2} & 81.7  \std{2.3} & 70.0  \std{3.6} & 62.3  \std{2.5} & 23.8  \std{1.2} & 26.6  \std{2.0} & 20.0  \std{1.4} & 22.1  \std{1.3} \\
        Fine Tuning & 72.7  \std{1.7} & 80.8  \std{2.0} & 68.7  \std{2.8} & 59.6  \std{3.1} & 22.1  \std{1.1} & 25.5  \std{1.5} & 18.1  \std{1.9} & 19.2  \std{1.6} \\
        MAML \citep{finn2017model} & 84.5  \std{1.7} & 97.3  \std{0.3} & 80.4  \std{0.3} & 63.5  \std{0.3} & 75.5  \std{0.7} & 88.8  \std{0.4} & 68.1  \std{0.5} & 56.2  \std{0.4} \\
        ANIL \citep{raghu2019rapid} & 75.3  \std{2.0} & 95.1  \std{0.6} & 58.7  \std{2.9} & 49.7  \std{0.3} & 69.1  \std{0.8} & 88.3  \std{0.5} & 52.4  \std{0.6} & 47.6  \std{0.9} \\
        BGD \citep{zeno2018task} & 87.8  \std{1.3} & 95.1  \std{0.5} & 86.9  \std{1.1} & 74.4  \std{1.1} & 63.4  \std{0.9} & 72.8  \std{1.2} & 55.9  \std{2.2} & 51.7  \std{1.3} \\
        MetaCOG  \citep{He2019TaskAC} & 88.0  \std{1.0} & 95.2  \std{0.5} & 87.1  \std{1.5} & 74.3  \std{1.5} & 63.6  \std{0.9} & 73.5  \std{1.3} & 55.9  \std{1.8} & 51.7  \std{1.4} \\
        MetaBGD  \citep{He2019TaskAC} & 91.1  \std{2.6} & 96.8  \std{1.5} & 92.5  \std{1.9} & 77.8  \std{3.8} & 74.8  \std{1.1} & 83.1  \std{1.0} & 71.7  \std{1.5} & 61.5  \std{1.2} \\
        \hline
        C-MAML & 89.5  \std{0.7} & 95.4  \std{0.4} & 91.1  \std{0.9} & 76.6  \std{1.3} & 82.6  \std{0.4} & 87.8  \std{0.4} & 84.6  \std{1.0} & 70.3  \std{0.7} \\
        C-MAML + KWTO & 92.2  \std{0.5} & 97.1  \std{0.3} &  94.1  \std{0.8} & 80.5  \std{1.4} & 84.5  \std{0.4} & 88.6  \std{0.5} & 86.2  \std{0.6} & 74.2  \std{0.8} \\
        C-MAML + KWTO + Acc. &  92.8  \std{0.6} &  97.8  \std{0.2} & 93.9  \std{0.8} & 79.9  \std{0.7} & 83.3  \std{0.4} & 89.0  \std{0.5} & 84.5  \std{0.7} & 71.1  \std{0.7} \\
    \end{tabular}
  \end{sc}
  \end{scriptsize}
  }
\caption{\textbf{Omniglot / MNIST / FashionMNIST experiment}}
% \vspace{-3mm}
\label{tab:omniglot}
\end{table*}

\subsection{Hyperparameter Sensitivity Analysis}
\label{app:hparam_analysis}

%% Using this repo:
%% https://app.wandb.ai/optimass/osaka_synbols_tbd0/

In this section, we analyze the \textit{update modulation} (UM) and \textit{prolonged adaptation phase} (PAP) mechanisms we introduce in C-MAML. Their respective hyperparameters are $\lambda$ and $\gamma$.

We perform the analysis on Synbols for the following reasons: (i) It is harder to solve than the Omniglot benchmark; (ii) Models train faster than Tiered-Imagenet; (iii) It is the only benchmark with an OoD task in which the pre-training data is bestowed a new semantic meaning, i.e., the font classification task.

We analyze the higher non-stationarity setting of $\alpha=0.98$. setting. This setting puts emphasis on  challenging the fundamental i.i.d assumption that CL is interested in solving.

\paragraph{Update Modulation}

\begin{figure}[h]
    \centering
    \includegraphics[width=0.8\textwidth]{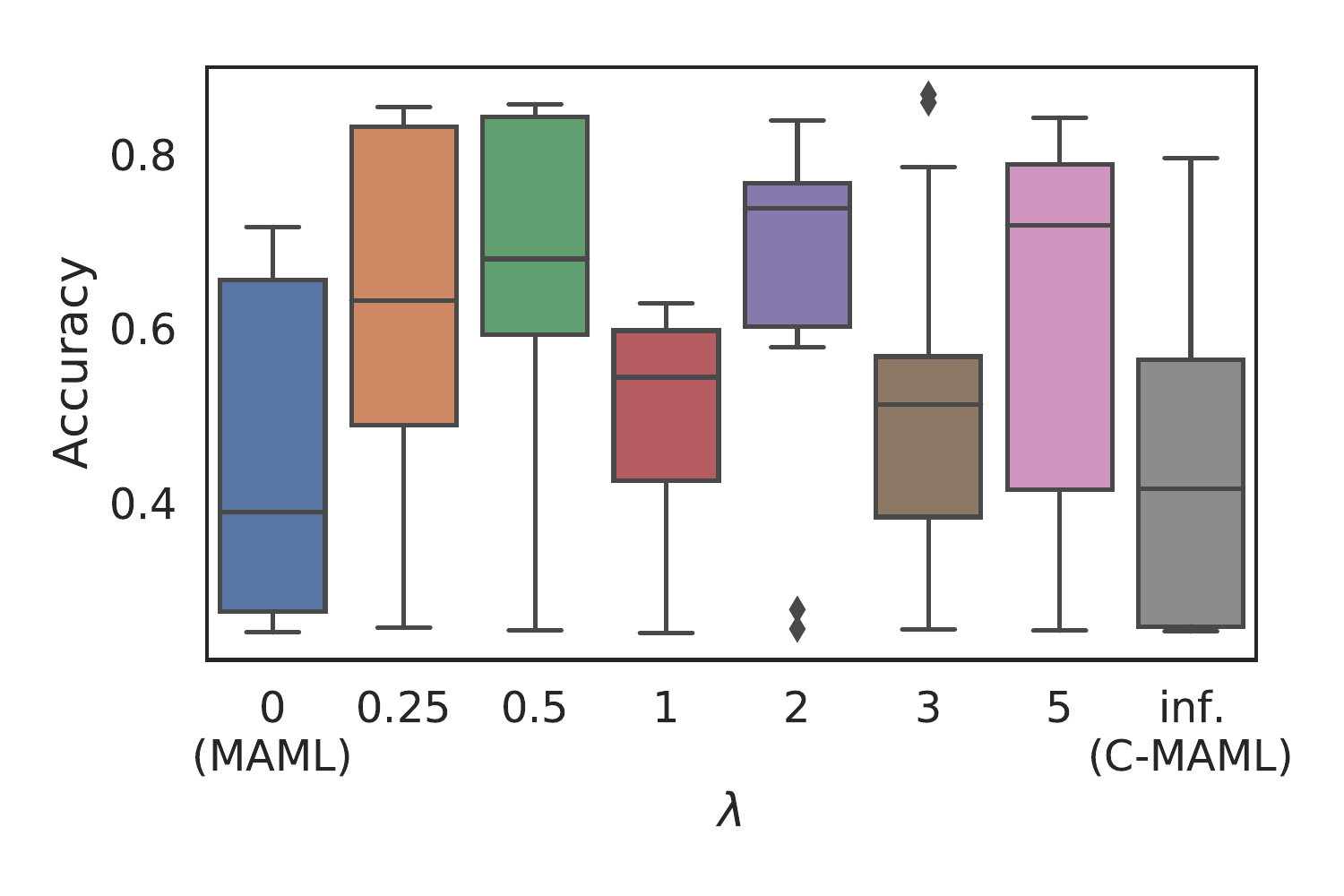}
    \caption{\textbf{Update modulation (UM) analysis.} The proposed mechanism is robust to its hyperparameter $\lambda$ and consistently increases average and maximum performance }
    \label{fig:lambda_analysis}
\end{figure}

We analyze the effect of $\lambda$ parameterizing $g_\lambda: \mathbb{R} \rightarrow (0,1)$. We use $g_\lambda$ to modulate the learning rate proportionally to the loss (see \cref{algo:cmaml}, L23). 
$\lambda$ provides a smooth interpolation between the behavior of MAML and Continual-MAML. When $\lambda=0$, Continual-MAML + UM collapses to MAML.  When $\lambda=\inf$, Continual-MAML + UM collapses to Continual-MAML. In Figure \ref{fig:lambda_analysis}, we show the effect of $\lambda$ on the online cumulative accuracy (same metric as reported elsewhere) which we obtained from our hyperparameter search. Interestingly, all values of $\lambda$ consistently increased the performance of Continual-MAML + UM with respect to MAML and Continual-MAML. This increase is due to two reasons. First, MAML ($\lambda=0$) cannot accumulate knowledge about the OoD tasks. Second, Continual-MAML (or $\lambda=\inf$) overfits its slow parameters $\phi$ to the current tasks, interfering with previous knowledge too aggressively. 
%UM is this robust to its hyperparameter.

\paragraph{Prolonged Adaptation Phase}

\begin{figure}
    \centering
    \includegraphics[width=0.4\textwidth]{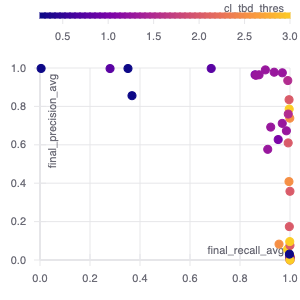}
    \includegraphics[width=0.4\textwidth]{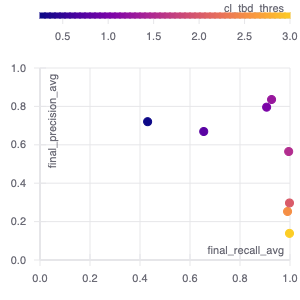}
    \caption{Precision (y-axis) and Recall (x-axis) for task boundary detection as a function of $\gamma$ (color). \textbf{Left:} all trials are plotted, \textbf{Right:} trials are grouped by $\gamma$ and the average is reported}
    \label{fig:pre_rec}
\end{figure}

To enable PAP, we need a mechanism to dectect the task boundary (or the context shifts). We propose a simple yet effective context shift detection mechanism which monitors the difference in loss with respect to the previous task and is controlled by a hyperparameter $\gamma$ (\cref{algo:cmaml}, L20). Setting $\gamma$ to high values will increase precision but reduce recall, and vice-versa. In Figure \ref{fig:pre_rec} we report precision and recall with respect to multiple values of $\gamma$. We can see that, when tuned appropriately, this mechanism can achieve near-perfect F1 scores, as highlighted by the trials near the top right corner. 

\begin{figure}[h] 
    \centering
    \includegraphics[width=0.8\textwidth]{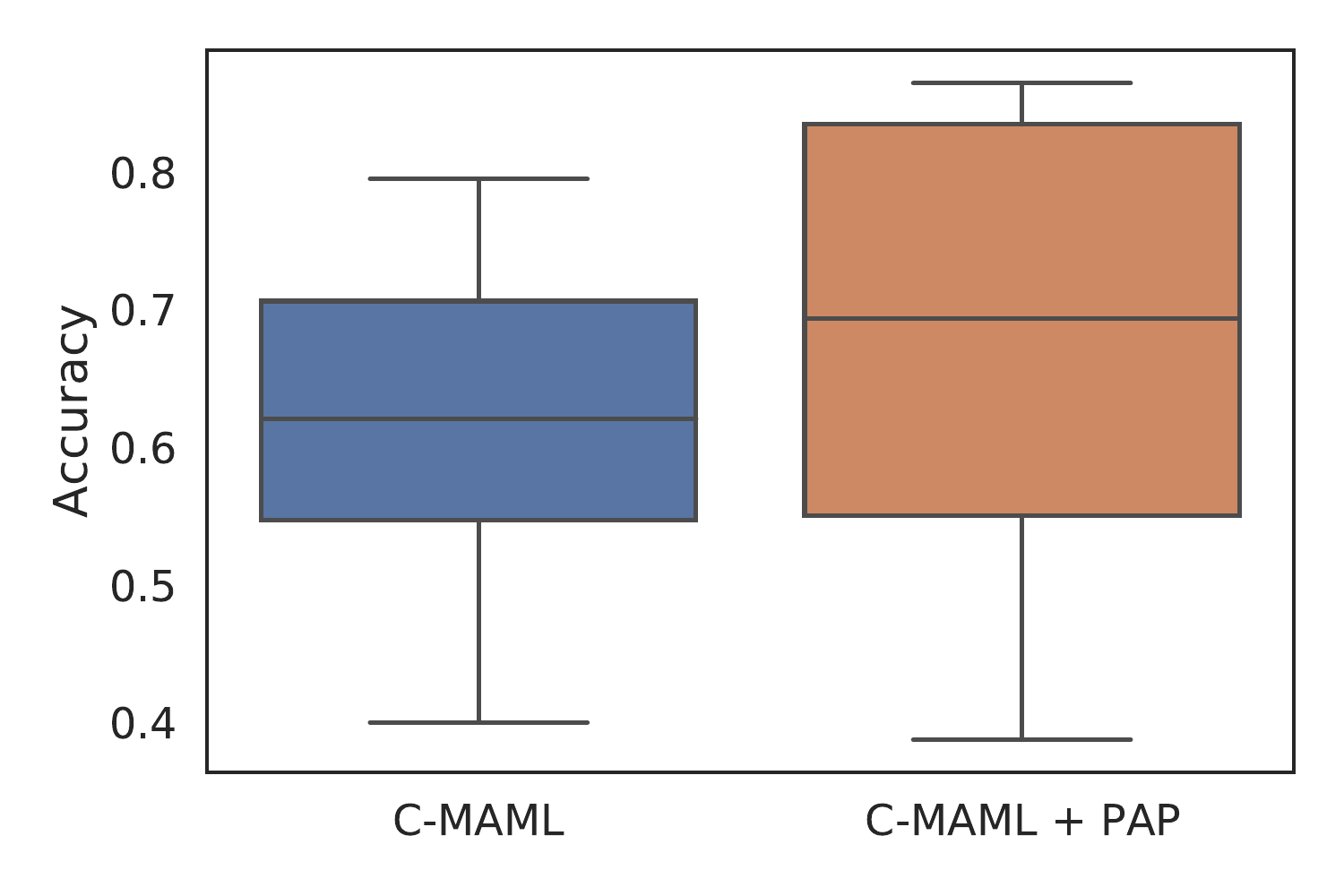}
    \caption{\textbf{Prolonged adaptation phase (PAP) analysis.} The proposed mechanism increases average and maximum performance. }
    \label{fig:PAP_analysis}
\end{figure}

The effectiveness of PAP is shown in Figure \ref{fig:PAP_analysis}. Specifically, we show that, across all values of $\gamma$, PAP increases the average performance of Continual-MAML. Again, the proposed mechanism is robust to its hyperparameter.

\clearpage
%-------------------------------------- FURTHER NOTES -----------------------------------------------------%
%----------------------------------------------------------------------------------------------------------%
\section{Extra Notes}
\label{app:further}

%  \subsection{Datasets}
 
%  The datasets used to construct the benchmarks are shown in Figure \ref{fig:datasets}

% % https://app.lucidchart.com/documents/edit/1c2566c0-b796-4f17-ba7c-74260d3c4115/_Trvls1L_HL7?shared=true&useCachedRole=false
% \begin{figure}[h]
%       \centering
%     \includegraphics[trim={3cm 3cm 3cm 3cm},clip,width=0.75\textwidth]{figures/omniglot2.png}
%     \includegraphics[width=0.75\textwidth]{figures/synbols3.png}
%     \includegraphics[trim={0.25cm 0.25cm 0.25cm 0.5cm},clip,width=0.75\textwidth]{figures/tiered.png}
%     \captionof{figure}{\textbf{Benchmarks.} We evaluate our setup on three different benchmarks, each one depicted in one row: Omniglot/MNIST/FashionMNIST (top), Synbols (middle), and Tiered-ImageNet (bottom).} 
%         \label{fig:datasets}
% \end{figure}

 \subsection{Bayesian Gradient Descent}
 \label{app:bgd}

Bayesian Gradient Descent (BGD) is a continual learning algorithm that models the distribution of the parameter vector $\phi$ by a factorized Gaussian. Similarly to \cite{He2019TaskAC} we apply BGD during the continual learning phase.
BGD models a the distribution of the parameter vector $\phi$ by a factorized Gaussian $q(\phi)=\prod_{i}\mathcal{N}(\phi_{i}|\mu_i, \sigma^2_i)$. Essential motivation behind BGD is that $\sigma$ models the uncertainty of the estimation of the parameter $\phi$. 
Hence parameters with higher uncertainty should be allowed to change faster than the parameters with lower $\sigma$, which are more important for preserving knowledge learned so far. BGD leverages variational Bayes techniques \citep{graves2011practical} and introduces an explicit closed-form update rule for the parameters $\mu_i$ and $\sigma_i$:
\begin{align*}
\mu_i = & \mu_i - \beta \sigma^2 \mathbb{E_\epsilon}(\frac{\partial \mathcal{L}\big(f_{\theta_{t-1}}(X_t), Y_t\big)}{\partial\phi}), \\
\sigma_i = & \sigma_i   \sqrt{1+(\frac{1}{2}\sigma_i\mathbb{E}_{\epsilon_{i}}[\frac{\partial \mathcal{L}\big(f_{\theta_{t-1}}(X_t), Y_t\big)}{\partial\phi_i}\epsilon_i])} -  \\
& \frac{1}{2}\sigma_i \mathbb{E}_{\epsilon_{i}}[\frac{\partial \mathcal{L}\big(f_{\theta_{t-1}}(X_t), Y_t\big)}{\partial\phi_i}\epsilon_i],
\end{align*} % to be corrected
where the expectations are approximated using Monte Carlo sampling and the re-parametrization trick is used as $\phi_i = \mu_i+\sigma_i\epsilon_i, \epsilon_i\sim\mathcal{N}(0,1)$.

\clearpage
%------------------------------------------------ Q&A -----------------------------------------------------%
%----------------------------------------------------------------------------------------------------------%
\section{Q\&A}
\label{app:QnA}

Here you can find reviewers questions and concerns and our answers that we couldn't address in the main part of the paper due to space limitation.

% \newcommand{\todo}[1]{\textcolor{red}{todo: #1}}
% \newcommand{\Rtwo}[1]{\textcolor{orange}{\bf R2}}
% \newcommand{\Rthree}[1]{\textcolor{cyan}{\bf R3}}
% \newcommand{\Rfour}[1]{\textcolor{ForestGreen}{\bf R4}}

% @\Rtwo{} % 2.1
% {\bf OSAKA seems too specific while failing to make a broader argument for other approaches to CL} 
% We wholeheartedly disagree. OSAKA's purpose is to align CL research with the \emph{deployment} of autonomous CL systems, e.g., a virtual assistant or a general-purpose robot that would keep on learning about new users and new environments (more examples in Sec. 1 \& 3). This scenario encompasses most of the reasons why we study CL. As hypothesised and then shown in the empirical section, other approaches to CL are not well suited to handle the broadness of real-life's requirements or of OSAKA's challenges and thus can not be applied at deployment time.  

% @\Rtwo{} % 2.2 and 2.7
{\bf Pre-training limits the generality of OSAKA and adds computational needs.}
We disagree. OSAKA aligns with the deployment of CL systems in real life (Sec. 1 \& 3) and it would be more realistic to deploy an agent with some knowledge of the world. Nevertheless, pre-training is not mandatory, although prescribed, and we have a baseline that does use it (C-MAML). Furthermore, it is currently more computationally efficient to learn on i.i.d. data at pre-training than on non-stationary data at CL time and pre-training is a one-time cost compared to CL which is a recurring one.

% @\Rthree{} % 3.1
{\bf Why putting features of different frameworks together is useful for continual learning evaluation?}
We unified \emph{and extended} these features to create a more realistic setting than the ones studied in the previous literature. Other frameworks study some of the features in silos but when methods are tested in less realistic settings some methods perform better than they should [12]. See Sec. 6.3 (under dynamic representations) for such an example.

% @\Rthree{} % 3.2
% {\bf several continual learning papers already measure the cumulative performance in task-agnostic setups, e.g. MOCA.}
% We have made a thorough literature review and we haven't found papers that measure cumulative performance apart from concurrent work MOCA. However, we are happy to add these papers if the reviewer includes them in their final review. Furthermore, Appendix B.1 is devoted to contrasting MOCA and OSAKA and describes their differences: context-dependant targets, OoD tasks and the expansion of the set of labels. \Rfour{} (and to some extent \Rtwo{}) agrees with the novelty of OSAKA as well as its potential to greatly enhance the field of CL.

% @\Rfour{} % 4.1
{\bf I think it is strange that MAML performs better in the 0.90 setting.}
The reviewer's intuition is right. However, C-MAML needs to predict correctly the context switches otherwise it will get mixed gradients from different tasks. Thus, $\alpha\!=\!0.90$ can be more challenging for methods with dynamic representations when the OoD tasks are not too far from the pre-training ones, as in the Tiered-Imagenet experiment.

{\bf Without task revisits, does $\phi$ stop being suitable for few-shot learning?} It stays suitable because it is still trained with the MAML loss, which optimizes for few-shot learning.